\documentclass{article} 
\usepackage{iclr2021_conference,times}

\usepackage{hyperref}
\usepackage{url}
\usepackage[utf8]{inputenc} 
\usepackage[T1]{fontenc}    
\usepackage{booktabs}       
\usepackage{amsmath}
\usepackage{amsthm}
\usepackage{amsfonts}       
\usepackage{nicefrac}       
\usepackage{microtype}      
\usepackage{todonotes}
\usepackage{cleveref}
\usepackage{multirow}
\usepackage{soul}
\usepackage{bm}
\usepackage{pgfplots}
\usepackage{caption}
\usepackage{algorithm}
\usepackage{algpseudocode}
\usepackage{xcolor,colortbl}
\usepackage{caption}
\usepackage{subcaption}
\usepackage{relsize}
\usepackage{ragged2e}

\algnewcommand{\LineComment}[1]{\State \(\triangleright\) #1}
\MakeRobust{\Call}

\newcommand{\tablefontsize}[0]{
\fontsize{7pt}{9pt}\selectfont
}

\newcommand{\tablefontsizeXS}[0]{
\fontsize{7.0pt}{9pt}\selectfont
}
\definecolor{Gray}{gray}{0.93}

\makeatletter
\newcommand{\printfnsymbol}[1]{%
  \textsuperscript{\@fnsymbol{#1}}%
}
\makeatother

\title{Degree-Quant: Quantization-Aware Training \\for Graph Neural Networks}

\author{Shyam A.~Tailor\thanks{Equal contribution. Correspondence to: Shyam Tailor <sat62@cam.ac.uk>} \\
  Department of Computer Science \& Technology \\
  University of Cambridge \\
  \And
  Javier Fernandez-Marques\printfnsymbol{1}\\
  Department of Computer Science \\
  University of Oxford \\
  \And 
  Nicholas D. Lane \\
  Department of Computer Science and Technology \\
  University of Cambridge \\
  \& Samsung AI Center 
}

\iclrfinalcopy 

\begin{document}

\maketitle

\begin{abstract}
Graph neural networks (GNNs) have demonstrated strong performance on a wide variety of tasks due to their ability to model non-uniform structured data.
Despite their promise, there exists little research exploring methods to make them more efficient at inference time.
In this work, we explore the viability of training quantized GNNs, 
enabling the usage of low precision integer arithmetic during inference.
For GNNs seemingly unimportant choices in quantization implementation cause dramatic changes in performance.
We identify the sources of error that uniquely arise when attempting to quantize GNNs, and propose an architecturally-agnostic and stable method, \emph{Degree-Quant}, to improve performance over existing quantization-aware training baselines commonly used on other architectures, such as CNNs. 
We validate our method on six datasets and show, unlike previous 
attempts, that models generalize to unseen graphs.
Models trained with Degree-Quant for INT8 quantization perform as well as FP32 models in most cases; for INT4 models, we obtain up to 26\% gains over the baselines. 
Our work enables up to 4.7$\times$ speedups on CPU when using INT8 arithmetic. 
\end{abstract}


\section{Introduction}

\begin{figure}[t]
\centering
\begin{minipage}{.48\textwidth}
  \centering
  \includegraphics[width=\linewidth,height=4.75cm,keepaspectratio]{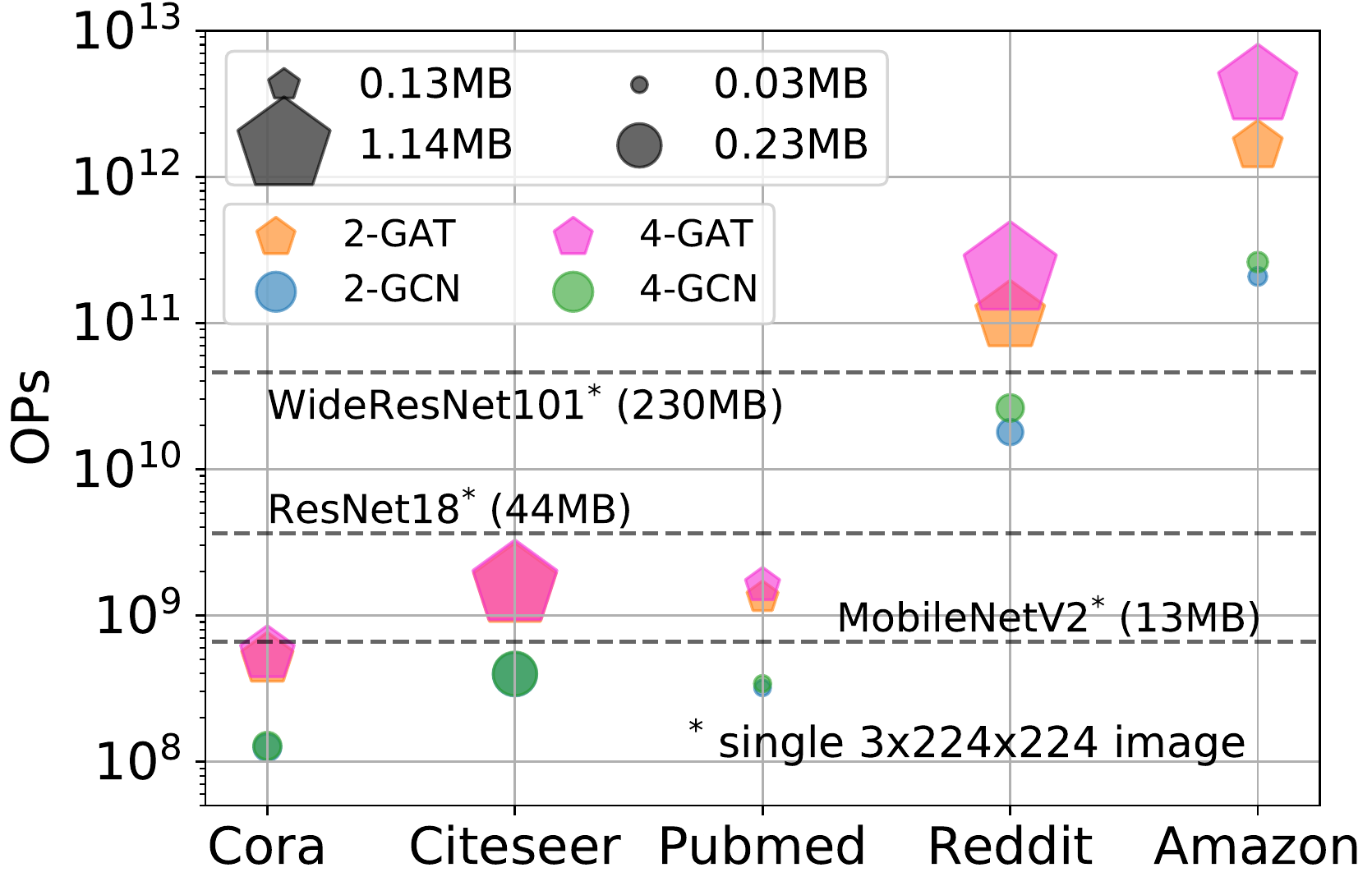}
  \captionsetup{font=small,labelfont=bf}
  \captionof{figure}{Despite GNN model sizes rarely exceeding 1MB, the OPs needed for inference grows at least linearly with the size of the dataset and node features. GNNs with models sizes $100\times$ smaller than popular CNNs require many more  OPs to process large graphs.
  }
  \label{fig:OPs}
\end{minipage}\hfill%
\begin{minipage}{.48\textwidth}
  \centering
  \newcommand{\sizein}{1.15}
\newcommand{\distout}{0.0125}
\newcommand{\distin}{0.5}
\newcommand{\gap}{0.125}
\newcommand{\opacity}{0.4}
\newcommand{\xDist}{3.0}
\newcommand{\yDist}{1.66} 

\newcommand{\angles}{60, 110, 220, 300}
\newcommand{\outangles}{0, 180}
\newcommand{\addangles}{0, 30, 330}

\definecolor{mygreen}{rgb}{0,0.3,0.6}
\definecolor{myred}{rgb}{0.8,0,0}

\usetikzlibrary{calc,patterns,positioning,decorations.pathmorphing,arrows}

\begin{tikzpicture}[
    box/.style = {draw, fill=teal, opacity=\opacity, minimum size=\sizein cm},  
    wbx/.style = {draw, fill=white, minimum size=\distin cm, outer sep=0pt},
    wbx2/.style = {draw, fill=white, minimum size=\distout cm, outer sep=0pt}
]

\foreach \i in {2,1,0}%
{
    \node   [box,above right] at (-\i*\gap,-\i*\gap*0.5) {};
    \node   [box,above right] at (\xDist-\i*\gap,-\i*\gap*0.5) {};
}
\node (w1) [wbx] at (0.3,0.3) {};
\node (w2) [wbx2] at (\xDist+0.3,0.3) {};
\draw[very thin]
    (w1.north west) -- (w2.north west)
    (w1.south east) -- (w2.south east);

\tikzstyle{vcenter} = [circle, draw, color=mygreen, fill=mygreen!5];
\tikzstyle{vadj} = [circle, draw, color=myred, fill=myred!5];
\tikzstyle{vgrey} = [circle, draw, color=gray, fill=gray!5];
\tikzstyle{eadj} = [draw, -stealth, very thick];
\tikzstyle{egrey} = [draw, -stealth, color=gray, opacity=0.5, decoration={snake, pre length=0.5mm, segment length=2mm, amplitude=0.5mm, post length=0.5mm}, decorate];

\node[vcenter, inner sep = 1pt] (h0) at (\sizein / 2, -\yDist) {\scriptsize $\mathbf{h}_v^l$};
\node[vcenter, inner sep = 1pt] (h1) at (\sizein / 2 + \xDist, -\yDist) {\scriptsize $\mathbf{h}_v^{l+1}$};

\foreach \a in \angles%
{
    \node[vadj] (\a0) at ([shift=({\a:0.8})]h0) {};
    \draw[eadj] (\a0) -- (h0);
    \foreach \b in \addangles%
    {
        \node[scale=0.1] (\a\b0) at ([shift=({\a + \b:0.5})]\a0) {};
        \draw[egrey] (\a\b0) -- (\a0);
    }
}

\foreach \a in \outangles%
{
    \node[vgrey] (\a0) at ([shift=({\a:0.6})]h0) {};
    \draw[egrey] (h0) -- (\a0);
}
\foreach \a in \angles%
{
    \node[vadj] (\a1) at ([shift=({\a:0.8})]h1) {};
    \draw[eadj] (\a1) -- (h1);
    \foreach \b in \addangles%
    {
        \node[scale=0.1] (\a\b1) at ([shift=({\a + \b:0.5})]\a1) {};
        \draw[egrey] (\a\b1) -- (\a1);
    }
}

\foreach \a in \outangles%
{
    \node[vgrey] (\a1) at ([shift=({\a:0.8})]h1) {};
    \draw[egrey] (h1) -- (\a1);
}

\foreach \a in \angles%
{
    \draw[dotted, thick] (\a0) -- (h1);
}

\node (cnntext) at (-1.25, \sizein / 2) {\footnotesize CNN};
\node (gnntext) at (-1.25, -\yDist) {\footnotesize GNN};

\node (cnneq) at (\xDist / 2 + \sizein / 2, -0.3) {\footnotesize $h^{l+1} = h^l \ast K$};
\node (gnneq) at (\xDist / 2 + \sizein / 2, -\yDist -1.4) {\footnotesize $\mathbf{h}_v^{l+1} = \gamma(\mathbf{h}_v^l, \bigwedge_{u \in \mathcal{N}(v)} [\phi(\mathbf{h}_u^l, \mathbf{h}_v^l, \mathbf{e}_{uv})])$};

\node (layer1) at (\sizein / 2, \sizein + 0.25) {\footnotesize Layer $l$};
\node (layer2) at (\sizein / 2 + \xDist, \sizein + 0.25) {\footnotesize Layer $l + 1$};

\end{tikzpicture}
  \captionsetup{font=small,labelfont=bf}
  \vspace*{-7mm}
  \captionof{figure}{While CNNs operate on regular grids, GNNs operate on graphs with varying topology.
  A node's neighborhood size and ordering varies for GNNs.
  Both architectures use weight sharing.}
\label{fig:convs}
\end{minipage}
\end{figure}

GNNs have received substantial attention in recent years due to their ability to model irregularly structured data. 
As a result, they are extensively used for applications as diverse as molecular interactions~\citep{duvenaud2015convolutional,wu2017moleculenet}, social networks~\citep{hamilton2017inductive}, recommendation systems~\citep{berg2017graph} or program understanding~\citep{allamanis2018learning}. 
Recent advancements have centered around building more sophisticated models
including new types of layers~\citep{kipf2017semi,velickovic2018graph,xu19how} and better aggregation functions~\citep{corso2020principal}. 
However, despite GNNs having few model parameters, the compute required for each application remains tightly coupled to the input graph size. 
A 2-layer Graph Convolutional Network (GCN) model with 32 hidden units would result in a model size of just 81KB but requires 19 GigaOPs to process the entire Reddit graph.
We illustrate this growth in \cref{fig:OPs}.





One major challenge with graph architectures is therefore performing inference efficiently, which limits the applications they can be deployed for.
For example, GNNs have been combined with CNNs for SLAM feature matching~\citep{sarlin2019superglue}, however it is not trivial to deploy this technique on smartphones, or even smaller devices, whose neural network accelerators often do not implement floating point arithmetic, and instead favour more efficient integer arithmetic.
Integer quantization is one way to lower the compute, memory and energy budget required to perform inference, without requiring modifications to the model architecture; this is also useful for model serving in data centers.



Although quantization has been well studied for CNNs and language models~\citep{jacob2017quantization, wang2018haq, zafrir2019q8bert, prato2019fully}, there remains relatively little work addressing GNN efficiency~\citep{10.1109/MICRO.2018.00010,mlsys2020_83, Zeng_2020, 10.1145/3394486.3403236}.
To the best of our knowledge, there is no work explicitly characterising the issues that arise when quantizing GNNs or showing latency benefits of using low-precision arithmetic. 
The recent work of \citet{wang2020binarized} explores only binarized embeddings of a single graph type (citation networks). 
In \citet{feng2020sgquant} a heterogeneous quantization framework assigns different bits to embedding and attention coefficients in each layer while maintaining the weights at full precision (FP32).
Due to the mismatch in operands' bit-width the majority of the operations are performed at FP32 after data casting, making it impractical to use in general purpose hardware such as CPUs or GPUs.
In addition they do not demonstrate how to train networks which generalize to \emph{unseen} input graphs.
Our framework relies upon uniform quantization applied to all elements in the network and uses bit-widths (8-bit and 4-bit) that are supported by off-the-shelf hardware such as CPUs and GPU for which efficient low-level operators for common operations found in GNNs exists.

This work considers the motivations and problems associated with quantization of graph architectures, and provides the following contributions:

\begin{itemize}
    \item The explanation of the sources of degradation in GNNs when using lower precision arithmetic. We show how the choice of straight-through estimator (STE) implementation, node degree, and method for tracking quantization statistics significantly impacts performance.
    
    \item An \emph{architecture-agnostic} method for quantization-aware training on graphs, \emph{Degree-Quant} (DQ), which results in INT8 models often performing as well as their FP32 counterparts. At INT4, models trained with DQ typically outperform quantized baselines by over 20\%.
    We show, unlike previous work, that models trained with DQ generalize to \emph{unseen graphs}.
    We provide code at this URL: \url{https://github.com/camlsys/degree-quant}.
        
    \item We show that quantized networks achieve up to 4.7$\times$ speedups on CPU with INT8 arithmetic, relative to full precision floating point, with 4-8$\times$ reductions in runtime memory usage.
        

\end{itemize}

\section{Background}
\subsection{Message Passing Neural Networks (MPNNs)}


Many popular GNN architectures may be viewed as generalizations of CNN architectures to an irregular domain: at a high level, graph architectures attempt to build representations based on a node's neighborhood (see \cref{fig:convs}).
Unlike CNNs, however, this neighborhood does not have a fixed ordering or size.
This work considers GNN architectures conforming to the MPNN paradigm~\citep{gilmer17neural}.
A graph $\mathcal{G} = (V, E)$ has node features $\mathbf{X} \in \mathbb{R}^{N \times F}$, an incidence matrix $\mathbf{I} \in \mathbb{N}^{2 \times E}$, and optionally $D$-dimensional edge features $\mathbf{E} \in \mathbb{R}^{E \times D}$.
The forward pass through an MPNN layer consists of message passing, aggregation and update phases: $\mathbf{h}_{l+1}^{(i)} = \gamma(\mathbf{h}_l^{(i)}, \bigwedge_{j \in \mathcal{N}(i)} [\phi(\mathbf{h}_l^{(j)}, \mathbf{h}_l^{(i)}, \mathbf{e}_{ij})])$.
Messages from node $u$ to node $v$ are calculated using function $\phi$, and are aggregated using a permutation-invariant function $\bigwedge$.
The features at $v$ are subsequently updated using $\gamma$.

We focus on three architectures with corresponding update rules:

\begin{enumerate}
    \setlength\itemsep{0em}
    \item Graph Convolution Network (GCN): $ \mathbf{h}_{l + 1}^{(i)} = \sum_{j \in \mathcal{N}(i) \cup \{i\}} ( \frac{1}{\sqrt{d_i d_j}} \mathbf{W} \mathbf{h}_l^{(j)} )$~\citep{kipf2017semi}, where $d_i$ refers to the degree of node $i$.
    
    \item Graph Attention Network (GAT): $\mathbf{h}_{l+1}^{(i)} = \alpha_{i,i}\mathbf{W}\mathbf{h}_l^{(i)} + \sum_{j \in \mathcal{N}(i)} ( \alpha_{i,j}\mathbf{W} \mathbf{h}_l^{(j)} )$, where $\alpha$ represent attention coefficients~\citep{velickovic2018graph}.
    
    \item Graph Isomorphism Network (GIN): $\mathbf{h}_{l+1}^{(i)} = f_{\mathbf{\Theta}}[ (1 + \epsilon) \mathbf{h}_l^{(i)} + \sum_{j \in \mathcal{N}(i)} \mathbf{h}_l^{(j)}]$, where $f$ is a learnable function (e.g.~a MLP) and $\epsilon$ is a learnable constant~\citep{xu19how}.
\end{enumerate}




\subsection{Quantization for Non-Graph Neural Networks}

Quantization allows for model size reduction and inference speedup without changing the model architecture.
While there exists extensive studies of the impact of quantization at different bit-widths~\citep{courbariaux2015binaryconnect,han2015deep,louizos2017bayesian} and data formats~\citep{micikevicius2017mixed,carmichael2018deep,kalamkar2019study}, it is 8-bit integer (INT8) quantization that has attracted the most attention.
This is due to INT8 models reaching comparable accuracy levels to FP32 models~\citep{krishnamoorthi2018quantizing, jacob2017quantization}, offer a $4\times$ model compression, and result in inference speedups on off-the-shelf hardware as 8-bit arithmetic is widely supported. 


Quantization-aware training (QAT) has become the \textit{de facto} approach towards designing robust quantized models with low error~\citep{wang2018haq,zafrir2019q8bert,wang2018haq}.
In their simplest forms, QAT schemes involve exposing the numerical errors introduced by quantization by simulating it on the forward pass \citet{jacob2017quantization} and make use of STE~\citep{bengio2013estimating} to compute the gradients---as if no quantization had been applied.
For integer QAT, the quantization of a tensor $x$ during the forward pass is often implemented as: $x_{q} = \min(q_{\max}, \max(q_{\min}, \lfloor x/s + z\rfloor))$, where $q_{\min}$ and $q_{\max}$ are the minimum and maximum representable values at a given bit-width and signedness, $s$ is the scaling factor making $x$ span the $[q_{\min}, q_{\max}]$ range and, $z$ is the \emph{zero-point}, which allows for the real value 0 to be representable in $x_{q}$. 
Both $s$ and $z$ are scalars obtained at training time. 
hen, the tensor is \emph{dequantized} as: $\hat{x} = (x_{q}-z)s$, where the resulting tensor $\hat{x} \sim x$ for a high enough bit-width. This similarity degrades at lower bit-widths. 
Other variants of integer QAT are presented in \citet{jacob2017quantization} and \citet{krishnamoorthi2018quantizing}.

To reach performance comparable to FP32 models, QAT schemes often rely on other techniques such as \textit{gradient clipping}, to mask gradient updates based on the largest representable value at a given bit-width; stochastic, or noisy, QAT, which stochastically applies QAT to a portion of the weights at each training step~\citep{fan2020training, dong2017learning}; or the re-ordering of layers~\citep{Sheng_2018, alizadeh2018a}. 

\section{Quantization for GNNs}
In this section, we build an intuition for why GNNs would fail with low precision arithmetic by identifying the sources of error that will disproportionately affect the accuracy of a low precision model.
Using this insight, we propose our technique for QAT with GNNs, \emph{Degree-Quant}.
Our analysis focuses on three models: GCN, GAT and GIN.
This choice was made as we believe that these are among the most popular graph architectures, with strong performance on a variety of tasks~\citep{dwivedi2020benchmarking}, while also being representative of different trends in the literature.

\subsection{Sources of Error}
\label{sec:sourcesOfError}

QAT relies upon the STE to make an estimate of the gradient despite the non-differentiable rounding operation in the forward pass.
If this approximation is inaccurate, however, then poor performance will be obtained.
In GNN layers, we identify the aggregation phase, where nodes combine messages from a varying number of neighbors in a permutation-invariant fashion, as a source of substantial numerical error, especially at nodes with high in-degree.
Outputs from aggregation have magnitudes that vary significantly depending on a node's in-degree: as it increases, the variance of aggregation values will increase.\footnote{The reader should note that we are not referring to the concept of estimator variance, which is the subject of sampling based approaches---we are exclusively discussing the variance of values immediately after aggregation.}
Over the course of training $q_{\min}$ and $q_{\max}$, the quantization range statistics, become severely distorted by infrequent outliers, reducing the resolution for the vast majority of values observed.
This reults in increased rounding error for nodes with smaller in-degrees.
Controlling $q_{\min}$ and $q_{\max}$ hence becomes a trade-off balancing \emph{truncation error} and \emph{rounding error}.

\begin{figure}
    \vspace{-5mm}
    \centering
    \includegraphics[width=0.82\textwidth]{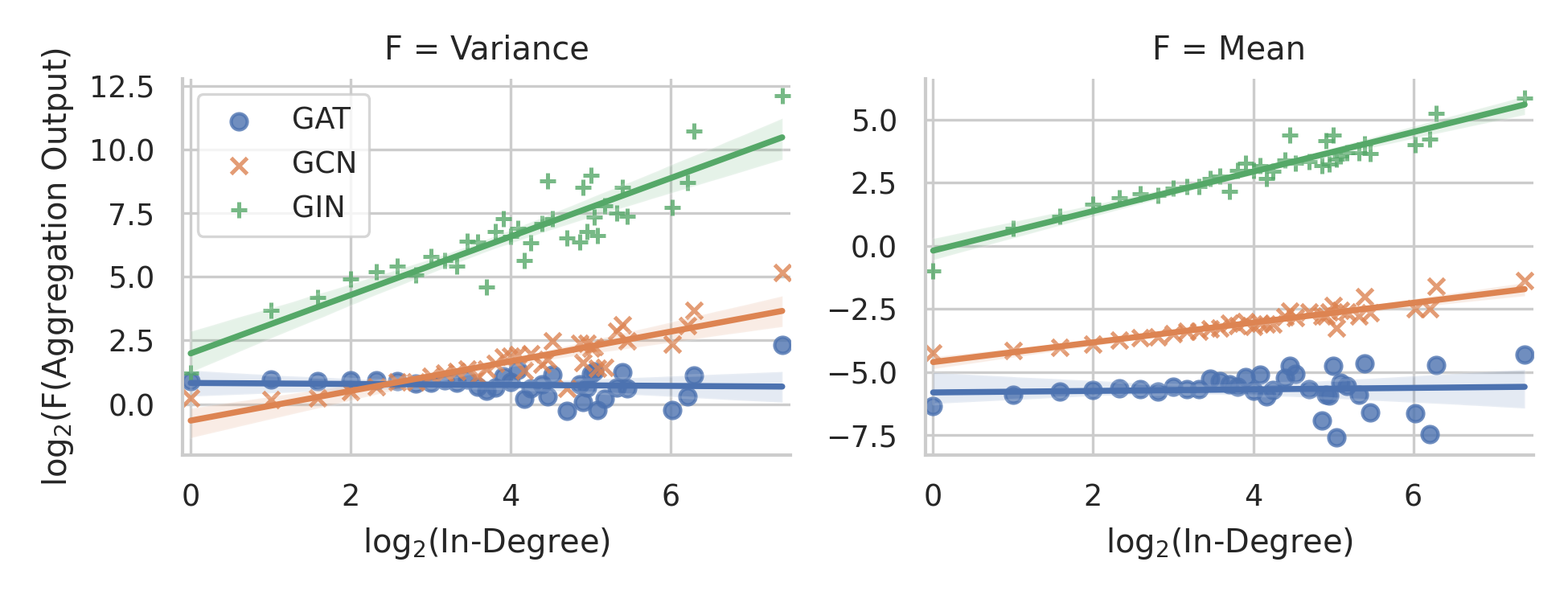}
    \captionsetup{font=small,labelfont=bf}
    \vspace{-4mm}
    \caption{Analysis of values collected immediately after aggregation at the final layer of FP32 GNNs trained on Cora.
    Generated using channel data collected from 100 runs for each architecture.
    As in-degree grows, so does the mean and variance of channel values after aggregation.}
    \label{fig:indegree}
\end{figure}

We can derive how the mean and variance of the aggregation output values vary as node in-degree, $n$, increases for each of the three GNN layers.
Suppose we model incoming message values for a single output dimension with random variables $X_i$, without making assumptions on their exact distribution or independence.
Further, we use $Y_n$ as the random variable representing the value of node output after the aggregation step.
With GIN layers, we have $Y_n = (1+ \epsilon)X_0 + \sum_{i=1}^n X_i$.
It is trivial to prove that $\mathbb{E}(Y_n) = \mathcal{O}(n)$.
The variance of the aggregation output is also $\mathcal{O}(n)$ in the case that that $\sum_{i \neq j}\text{Cov}(X_i, X_j) \ll \sum_i \text{Var}(X_i)$.
We note that if $\sum_{i\neq j}\text{Cov}(X_i, X_j)$ is large then it implies that the network has learned highly redundant features, and may be a sign of over-fitting.
Similar arguments can be made for GCN and GAT layers; we would expect GCN aggregation values to grow like $\mathcal{O}(\sqrt{n})$, and GAT aggregation values to remain constant ($\mathcal{O}(1)$) due to the attention coefficients.

We empirically validate these predictions on GNNs trained on Cora; results are plotted in \cref{fig:indegree}.
We see that the aggregation values do follow the trends predicted, and that for the values of in-degree in the plot (up to 168) the covariance terms can be neglected.
As expected, the variance and mean of the aggregated output grow fastest for GIN, and are roughly constant for GAT as in-degree increases.
From this empirical evidence, it would be expected that GIN layers are most affected by quantization.

By using GIN and GCN as examples, we can see how aggregation error causes error in weight updates.
Suppose we consider a GIN layer incorporating one weight matrix in the update function i.e.~$\mathbf{h}^{(i)}_{l+1} = f(\mathbf{W}\mathbf{y}^{(i)}_{\text{GIN}})$, where $f$ is an activation function, $\mathbf{y}^{(i)}_{\text{GIN}} = (1 + \epsilon)\mathbf{h}^{(i)}_{l} + \sum_{j \in \mathcal{N}(i)} \mathbf{h}^{(j)}_{l}$, and $\mathcal{N}(i)$ denotes the in-neighbors of node $i$.
Writing $\mathbf{y}^{(i)}_{\text{GCN}} = \sum_{k \in \mathcal{N}(i)} ( \frac{1}{\sqrt{d_i d_k}} \mathbf{W}\mathbf{h}_l^{(j)} )$, we see that the derivatives of the loss with respect to the weights for GCN and GIN are:

\noindent\begin{minipage}{.4\linewidth}
\centering
\textbf{GIN}
\vspace{-2mm}
\footnotesize
$$\frac{\partial \mathcal{L}}{\partial \mathbf{W}} = \mathlarger{\sum}_{i=1}^{|V|}\left( \frac{\partial \mathcal{L}}{\partial \mathbf{h}^{(i)}_{l+1}} \circ f'(\mathbf{W}\mathbf{y}^{(i)}_{\text{GIN}}) \right) \mathbf{y}_{\text{GIN}}^{(i)^\top} $$
\end{minipage}\hfill%
\begin{minipage}{.55\linewidth}
\centering
\textbf{GCN}
\vspace{-2mm}
\footnotesize
$$\frac{\partial \mathcal{L}}{\partial \mathbf{W}} = \sum_{i=1}^{|V|} \sum_{j \in \mathcal{N}(i)} \frac{1}{\sqrt{d_i d_j}} \left( \frac{\partial \mathcal{L}}{\partial \mathbf{h}^{(i)}_{l+1}} \circ f'(\mathbf{y}^{(i)}_{\text{GCN}}) \right) \mathbf{h}_l^{(j)^\top}$$
\end{minipage}

The larger the error in $\mathbf{y}^{(i)}_{\text{GIN}}$---caused by aggregation error---the greater the error in the weight gradients for GIN, which results in poorly performing models being obtained.
The same argument applies to GCN, with the $\mathbf{h}_l^{(j)^\top}$ and $\mathbf{y}^{(i)}_{\text{GCN}}$ terms introducing aggregation error into the weight updates.

\subsection{Our Method: Degree-Quant}
\label{sec:dq}
\begin{figure}
    \centering
    \usetikzlibrary{arrows,shapes}

\definecolor{mygreen}{rgb}{0,0.3,0.6}
\definecolor{myred}{rgb}{0.8,0,0}

\newcommand{\points}{1.4/1.2/a/vlo, 2.5/1.6/b/vhi, 4/1.6/c/vlo, 1.8/0.5/d/vlo, 3/1/e/vhi, 3/0.2/f/vlo, 4/0.5/g/vlo}
\newcommand{\edges}{a/b/elo, c/b/elo, d/a/elo, d/b/elo, b/e/ehi, e/c/ehi, d/e/elo, f/d/elo, f/e/elo, g/e/elo, g/f/elo}

\pgfdeclarelayer{background}
\pgfsetlayers{background,main}

\tikzstyle{vhi} = [circle, draw, color=mygreen, fill=mygreen!5]
\tikzstyle{vlo} = [rectangle, draw, color=myred, fill=myred!5]
\tikzstyle{ehi} = [draw, -stealth, color=mygreen, thick]
\tikzstyle{elo} = [draw, -stealth, color=myred, thick]

\tikzstyle{stage} = [draw, -stealth, thick]

\begin{tikzpicture}
\foreach \x/\y/\name/\style in \points
    \node[vhi] (\name0) at (\x, \y) {};

\foreach \x/\y/\name/\style in \points
    \node[\style] (\name1) at (\x + 5.25, \y) {};

\foreach \source/ \dest /\style in \edges
    \draw[ehi] (\source0) -- (\dest0);

\foreach \source/ \dest /\style in \edges
    \draw[\style] (\source1) -- (\dest1);
    
\node (s1) at (4.7, 0.9) {};
\node (s2) at (6, 0.9) {};
\draw[stage] (s1) -- node[anchor=south, align=center] {\footnotesize Protect \\ \footnotesize \& Quantize} (s2);

\node (s3) at (9.75, 0.9) {};
\node (s4) at (11.7, 0.9) {\Large \hspace{2mm} \ldots};
\draw[stage] (s3) -- node[anchor=south, align=center] {\footnotesize Aggregate \\ \footnotesize \& Update} (s4);

\node (s00) at (-0.75, 0.9) {\Large \ldots};
\node (s01) at (1, 0.9) {};
\draw[stage] (s00) -- (s01);
\end{tikzpicture} 
    \captionsetup{font=small,labelfont=bf}
    \caption{High-level view of the stochastic element of Degree-Quant.
    Protected (high in-degree) nodes, in blue, operate at full precision, while unprotected nodes (red) operate at reduced precision.
    High in-degree nodes contribute most to poor gradient estimates, hence they are stochastically protected from quantization more often.}
    \label{fig:degree_quant}
\end{figure}
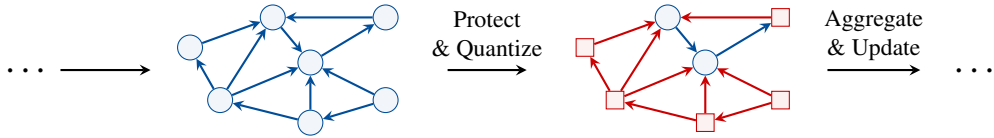

To address these sources of error we propose \emph{Degree-Quant} (DQ), a method for QAT with GNNs.
We consider both \emph{inaccurate weight updates} and \emph{unrepresentative quantization ranges}.

\begin{algorithm}
\captionsetup{font=small,labelfont=bf}
\caption{Degree-Quant (DQ). Functions accepting a protective mask $\mathbf{m}$ perform only the masked computations at full precision: intermediate tensors are \emph{not} quantized. At test time protective masking is disabled. In \cref{fig:diagram_dq} (in the Appendix) we show with a diagram how a GCN layers makes use of DQ.}\label{alg:dq_pseudo}
\small
\begin{algorithmic}[1]
\Procedure{TrainForwardPass}{$\mathcal{G}, \mathbf{p}$}
\LineComment{Calculate mask and \emph{quantized} weights, $\Theta'$, which all operations share}
\State $\mathbf{m}\gets$ \Call{Bernoulli}{$\mathbf{p}$}
\State $\Theta'\gets$ \Call{Quantize}{$\Theta$}
\LineComment{Messages with masked sources are at full precision (excluding weights)}
\State $\mathcal{M}\gets$\Call{MessageCalculate}{$\mathcal{G}, \Theta', \mathbf{m}$}
\State $X\gets$ \Call{Quantize}{\Call{Aggregate}{$\mathcal{M}, \Theta', \mathbf{m}$}, $\mathbf{m}$} \Comment{No quantization for masked nodes}
\State \Return \Call{Update}{$X, \Theta', \mathbf{m}$}\Comment{Quantized weights always used}
\EndProcedure
\end{algorithmic}
\end{algorithm}

\textbf{Stochastic Protection from Quantization to Improve Weight Update Accuracy}.
DQ aims to encourage more accurate weight updates by stochastically protecting nodes in the network from quantization.
At each layer a protective node mask is generated; all masked nodes have the phases of the message passing, aggregation and update performed at full precision.
This includes messages sent by protected nodes to other nodes, as shown in \cref{fig:degree_quant} (a detailed diagram is shown in \cref{fig:diagram_dq}).
It is also important to note that the weights used at all nodes are the same quantized weights; this is motivated by the fact that our method is used to encourage more accurate gradients to flow back to the weights through high in-degree nodes.
At test time protection is disabled: all nodes operate at low precision.

To generate the mask, we pre-process each graph before training and create a vector of probabilities $\mathbf{p}$ with length equal to the number of nodes.
At training time, mask $\mathbf{m}$ is generated by sampling using the Bernoulli distribution: $\mathbf{m} \sim \text{Bernoulli}(\mathbf{p})$.
In our scheme $p_i$ is higher if the in-degree of node $i$ is large, as we find empirically that high in-degree nodes contribute most towards error in weight updates.
We use a scheme with two hyperparameters, $p_{\min}$ and $p_{\max}$; nodes with the maximum in-degree are assigned $p_{\max}$ as their masking probability, with all other nodes assigned a probability calculated by interpolating between $p_{\min}$ and $p_{\max}$ based on their in-degree ranking in the graph.

\textbf{Percentile Tracking of Quantization Ranges}.
\Cref{fig:indegree} demonstrates large fluctuations in the variance of the aggregation output as in-degree increases.
Since these can disproportionately affect the ranges found by using min-max or momentum-based quantization, we propose using \emph{percentiles}.
While percentiles have been used for post-training quantization ~\citep{wu2020integer}, we are the first (to the best of our knowledge) to propose making it a core part of QAT; we find it to be a key contributor to achieving consistent results with graphs.
Using percentiles involves ordering the values in the tensor and clipping a fraction of the values at both ends of the distribution.
The fraction to clip is a hyperparameter.
We are more aggressive than existing literature on the quantity we discard: we clip the top and bottom 0.1\%, rather than 0.01\%, as we observe the fluctuations to be a larger issue with GNNs than with CNNs or DNNs.
Quantization ranges are more representative of the vast majority of values in this scheme, resulting in less \emph{rounding error}.

We emphasize that a core contribution of DQ is that it is \emph{architecture-agnostic}.
Our method enables a wide variety of architectures to use low precision arithmetic at inference time.
Our method is also \emph{orthogonal}---and complementary---to other techniques for decreasing GNN computation requirements, such as sampling based methods which are used to reduce memory consumption~\citep{graphsaint}, or weight pruning~\citep{blalock2020state} approaches to achieve further model compression.

\section{Experiments}


\label{sec:experiments}
In this section we first analyse how the choice of quantization implementation affects performance of GNNs.
We subsequently evaluate Degree-Quant against the strong baselines of: FP32, INT8-QAT and, INT8-QAT with stochastic masking of weights~\citep{fan2020training}. We refer to this last approach as \textit{noisy} QAT or nQAT.
To make explicit that we are quantizing both weights and activations, we use the notation W8A8. 
We repeat the experiments at INT4. 
Our study evaluates performance on six datasets and includes both node-level and graph-level tasks. 
The datasets used were Cora, CiteSeer, ZINC, MNIST and CIFAR10 superpixels, and REDDIT-BINARY.
Across all datasets INT8 models trained with Degree-Quant manage to recover most of the accuracy lost as a result of quantization. In some instances, DQ-INT8 outperform the extensively tuned FP32 baselines. For INT4, DQ outperforms all QAT baselines and results in double digits improvements over QAT-INT4 in some settings.
Details about each dataset and our experimental setup can be found in \cref{app:experimentalSetup}.

\begin{table*}[t]
    \centering
    \resizebox{\textwidth}{!}{%
    \begin{tabular}{p{1.1cm} c | c c | c c | c c | c c}
        \toprule
         &  &\multicolumn{4}{c|}{\textit{vanilla} STE} & \multicolumn{4}{c}{STE with Gradient Clipping} \\
         Dataset & Model & \multicolumn{2}{c}{min/max} & \multicolumn{2}{c|}{momentum} & \multicolumn{2}{c}{min/max} & \multicolumn{2}{c}{momentum} \\
         & Arch.& W8A8 & W4A4 & W8A8 & W4A4 & W8A8 & W4A4 & W8A8 & W4A4 \\
        \midrule
        \multirow{3}{*}{\parbox{1.1cm}{Cora (Acc.~\%)~$\uparrow$}} & GCN & $\bm{81.0\pm0.7}$ & $65.3\pm4.9$ & $42.3\pm11.1$ & $49.4\pm8.8$ & $80.8\pm0.8$ & $62.3\pm5.2$ & $66.9\pm18.2$ & \bm{$77.2\pm2.5$} \\
         & GAT & $76.0\pm2.2$ & $16.8\pm8.5$ & $81.7\pm1.3$ & \bm{$51.7\pm5.8$} & $76.4\pm2.6$ & $15.4\pm8.1$ & $\bm{81.9\pm0.7}$ & $47.4\pm5.0$ \\
         & GIN & $69.9\pm1.9$ & $ 25.9\pm2.6$ & $49.2\pm10.2$ & \bm{$42.8\pm4.0$} & $69.2\pm2.3$ & $29.5\pm3.5$ & $\bm{75.1\pm1.1}$ & $40.5\pm5.0$ \\
        \midrule
        \multirow{3}{*}{\parbox{1.1cm}{MNIST (Acc.~\%)~$\uparrow$}} & GCN & $\bm{90.4\pm0.2}$ & $51.3\pm7.5$ & $90.1\pm0.5$ & \bm{$70.6\pm2.4$} & $90.4\pm0.3$ & $54.8\pm1.5$ & $90.2\pm0.4$ & $10.3\pm0.0$ \\
         & GAT & $\bm{95.8\pm0.1}$ & $20.1\pm3.3$ & $95.7\pm0.3$ & $67.4\pm3.2$ & $95.7\pm0.1$ & $30.2\pm7.4$ & $95.7\pm0.3$ & \bm{$76.3\pm1.2$} \\
         & GIN & $96.5\pm0.3$ & $62.4\pm21.8$ & \bm{$96.7\pm0.2$} & \bm{$91.0\pm0.6$} & $96.4\pm0.4$ & $19.5\pm2.1$ & $75.3\pm18.1$ & $10.8\pm0.9$ \\
        \midrule
        \multirow{3}{*}{\parbox{1.1cm}{ZINC (Loss)~$\downarrow$}} & GCN & $0.486\pm0.01$ & $0.747\pm0.02$ & $0.509\pm0.01$ & $0.710\pm0.05 $ & $0.495\pm0.01$ & $0.766\pm0.02$ & $\bm{0.483\pm0.01}$ & \bm{$0.692\pm0.01$} \\
         & GAT & $0.471\pm0.01$ & $0.740\pm0.02$ & $0.571\pm0.03$ & \bm{$0.692\pm0.06$} & $0.466\pm0.01$ & $0.759\pm0.04$ & $\bm{0.463\pm0.01}$ & $0.717\pm0.03$ \\
         & GIN & $0.393\pm0.02$ & $1.206\pm0.27$ & \bm{$0.386\pm0.03$} & \bm{$0.572\pm0.02$} & $0.390\pm0.02$ & $1.669\pm0.10$ & $0.388\pm0.02$ & $0.973\pm0.24$ \\
        \bottomrule
    \end{tabular}%
    }
    \captionsetup{font=small,labelfont=bf}
    \caption{
    Impact on performance of four typical quantization implementations for INT8 and INT4.
    The configuration that resulted in best performing models for each dataset-model pair is bolded.
    Hyperparameters for each experiment were fine-tuned independently.
    As expected, adding clipping does not change performance with min/max but does with momentum.
    \textbf{A major contribution of this work is identifying that seemingly unimportant choices in quantization implementation cause dramatic changes in performance.}
    }
    \label{tab:initQuant}
\end{table*}

\subsection{Impact of Quantization Gradient Estimator on Convergence}
\label{sec:STEconfig}


The STE is a workaround for when the forward pass contains non-differentiable operations (e.g.~rounding in QAT) that has been widely adopted in practice.
While the choice of STE implementation generally results in marginal differences for CNNs---even for binary networks~\citep{alizadeh2018a}---it is unclear whether only marginal differences will also be observed for GNNs. 
Motivated by this, we study the impact of four off-the-shelve quantization procedures on the three architectures evaluated for each type of dataset; the implementation details of each one is described in \cref{app:STEconfig}.
We perform this experiment to ensure that we have the strongest possible QAT baselines.
Results are shown in \cref{tab:initQuant}. We found the choice quantization implementation to be highly dependent on the model architecture and type of problem to be solved: we see a much larger variance than is observed with CNNs; this is an important discovery for future work building on our study.

We observe a general trend in all INT4 experiments benefiting from momentum as it helps smoothing out the quantization statistics for the inherently noisy training stage at low bitwidths.
This trend applies as well for the majority of INT8 experiments, while exhibiting little impact on MNIST.
For INT8 Cora-GCN, large gradient norm values in the early stages of training (see \cref{fig:norm_min_max}) mean that these models not benefit from momentum as quantization ranges fail to keep up with the rate of changes in tensor values; higher momentum can help but also leads to instability.
In contrast, GAT has stable initial training dynamics, and hence obtains better results with momentum.
For the molecules dataset ZINC, we consistently obtained lower regression loss when using momentum.
We note that GIN models often suffer from higher performance degradation (as was first noted in \cref{fig:indegree}), specially at W4A4.
This is not the case however for image datasets using superpixels. 
We believe that datasets with Gaussian-like node degree distributions (see \cref{fig:indegree_dist}) are more tolerant of the imprecision introduced by quantization, compared to datasets with tailed distributions.
We leave more in-depth analysis of how graph topology affects quantization as future work.



\begin{table*}[t]
    \tablefontsize
    \setlength{\tabcolsep}{4pt}
    \centering
    \resizebox{\linewidth}{!}{
    \begin{tabular}{p{0.7cm} c c c c c c}
        \toprule
        Quant. & Model & \multicolumn{2}{c}{\tablefontsizeXS Node Classification (Accuracy \%)} & \multicolumn{2}{c}{\tablefontsizeXS Graph Classification (Accuracy \%)} & \multicolumn{1}{c}{\tablefontsizeXS Graph Regression (Loss)} \\
        
        Scheme & Arch. & Cora $\uparrow$ & Citeseer $\uparrow$ & MNIST $\uparrow$ & CIFAR-10 $\uparrow$ & ZINC $\downarrow$ \\
        \midrule
        \multirow{3}{*}{\parbox{0.7cm}{Ref. (FP32)}} & GCN & $81.4\pm0.7$ & $71.1\pm0.7$ & $90.0\pm0.2$ & $54.5\pm0.1$ & $0.469\pm0.002$ \\
         & GAT & $83.1\pm0.4$ & $72.5\pm0.7$ &  $95.6\pm0.1$ & $65.4\pm0.4$ & $0.463\pm0.002$ \\
         & GIN & $77.6\pm1.1$ & $66.1\pm0.9$ & $93.9\pm0.6$ & $53.3\pm3.7$ & $0.414\pm0.009$ \\
        \midrule
        \multirow{3}{*}{\parbox{0.7cm}{Ours (FP32)}} & GCN & $81.2\pm0.6$ & $71.4\pm0.9$ & $90.9\pm0.4$ & $58.4\pm0.5$ & $0.450\pm0.008$ \\
         & GAT & $83.2\pm0.3$ & $72.4\pm0.8$ & $95.8\pm0.4$ & $65.1\pm0.8$ & $0.455\pm0.006$ \\
         & GIN & $77.9\pm1.1$ & $65.8\pm1.5$ & $96.4\pm0.4$ & $57.4\pm0.7$ & $0.334\pm0.024$ \\
        \midrule
        \midrule
        \multirow{3}{*}{\parbox{0.7cm}{QAT (W8A8)}} & GCN & $81.0\pm0.7$ & $71.3\pm1.0$ & $90.9\pm0.2$ & $56.4\pm0.5$ & $0.481\pm0.029$ \\
         & GAT & $81.9\pm0.7$ & $71.2\pm1.0$ & $95.8\pm0.3$ & $66.3\pm0.4$ & $0.460\pm0.005$ \\
         & GIN & $75.6\pm1.2$ & $63.0\pm2.6$ & $96.7\pm0.2$ & $52.4\pm1.2$ & $0.386\pm0.025$ \\
        \midrule
        \multirow{3}{*}{\parbox{0.7cm}{nQAT (W8A8)}} & GCN & $81.0\pm0.8$ & $70.7\pm0.8$ & $91.1\pm0.1$ & $56.2\pm0.5$ & $0.472\pm0.015$ \\
         & GAT & $82.5\pm0.5$ & $71.2\pm0.7$ & $96.0\pm0.1$ & $66.7\pm0.2$ & $0.459\pm0.007$ \\
         & GIN & $77.4\pm1.3$ & $65.1\pm1.4$ & $96.4\pm0.3$ & $52.7\pm1.4$ & $0.405\pm0.016$ \\
        \midrule
        \rowcolor{Gray}
         & GCN & \tablefontsizeXS $81.7\pm0.7$ (+$\bm{0.7}$) & \tablefontsizeXS $71.0\pm0.9$ (-$\bm{0.3}$) & \tablefontsizeXS $90.9\pm0.2$ (-$\bm{0.2}$) & \tablefontsizeXS $56.3\pm0.1$ (-$\bm{0.1}$) & \tablefontsizeXS $0.434\pm0.009$ (+$\bm{9.8}$) \\
         \rowcolor{Gray}
         & GAT & \tablefontsizeXS$82.7\pm0.7$ (+$\bm{0.2}$) & \tablefontsizeXS$71.6\pm1.0$ (+$\bm{0.4}$) & \tablefontsizeXS $95.8\pm0.4$ (-$\bm{0.2}$) & \tablefontsizeXS$67.7\pm0.5$ (+$\bm{1.0}$) & \tablefontsizeXS$0.456\pm0.005$ (+$\bm{0.9}$) \\
         \rowcolor{Gray}
         \multirow{-3}{*}{\parbox{0.7cm}{DQ (W8A8)}}& GIN & \tablefontsizeXS$78.7\pm1.4$ (+$\bm{1.3}$) &  \tablefontsizeXS$67.5\pm1.4$ (+$\bm{2.4}$) & \tablefontsizeXS $96.6\pm0.1$ (-$\bm{0.1}$) & \tablefontsizeXS$55.5\pm0.6$ (+$\bm{2.8}$) & \tablefontsizeXS$0.357\pm0.014$ (+$\bm{7.5}$) \\
        \midrule
        \midrule
        \multirow{3}{*}{\parbox{0.7cm}{QAT (W4A4)}} & GCN & $77.2\pm2.5$ & $64.1\pm4.1$ & $70.6\pm2.4$ & $38.1\pm1.6$ & $0.692\pm0.013$ \\
         & GAT & $55.6\pm5.4$ & $65.3\pm1.9$ & $76.3\pm1.2$ & $41.0\pm1.1$ & $0.655\pm0.032$ \\
         & GIN & $42.5\pm4.5$ & $18.6\pm2.9$ & $91.0\pm0.6$ & $45.6\pm3.6$ & $0.572\pm0.02$ \\
        \midrule
        \multirow{3}{*}{\parbox{0.7cm}{nQAT (W4A4)}} & GCN & $78.1\pm1.5$ & $65.8\pm2.6$ & $70.9\pm1.5$ & $40.1\pm0.7$ & $0.669\pm0.128$ \\
         & GAT & $54.9\pm5.6$ & $65.5\pm1.7$ & $78.4\pm1.5$ & $41.0\pm0.6$ & $0.637\pm0.012$ \\
         & GIN & $45.0\pm5.0$ & $34.6\pm3.8$ & $91.3\pm0.5$ & $48.7\pm1.7$ & $0.561\pm0.068$ \\
        \midrule
        \rowcolor{Gray}
         & GCN & $78.3\pm1.7$ (+$\bm{0.2}$) & \tablefontsizeXS $66.9\pm2.4$ (+$\bm{1.1}$) & \tablefontsizeXS $84.4\pm1.3$ (+$\bm{13.5}$) & \tablefontsizeXS $51.1\pm0.7$ (+$\bm{11.0}$) & \tablefontsizeXS $0.536\pm0.011$ (+$\bm{26.2}$) \\
         \rowcolor{Gray}
         & GAT & $71.2\pm2.9$ (+$\bm{16.3}$) & \tablefontsizeXS $67.6\pm1.5$ (+$\bm{2.1}$) & \tablefontsizeXS $93.1\pm0.3$ (+$\bm{14.7}$) & \tablefontsizeXS $56.5\pm0.6$ (+$\bm{15.5}$) & \tablefontsizeXS $0.520\pm0.021$ (+$\bm{20.6}$) \\
         \rowcolor{Gray}
         \multirow{-3}{*}{\parbox{0.7cm}{DQ (W4A4)}} & GIN & $69.9\pm3.4$ (+$\bm{24.9}$) & \tablefontsizeXS $60.8\pm2.1$ (+$\bm{26.2}$) & \tablefontsizeXS $95.5\pm0.4$ (+$\bm{4.2}$) & \tablefontsizeXS $50.7\pm1.6$ (+$\bm{2.0}$) & \tablefontsizeXS $0.431\pm0.012$ (+$\bm{23.2}$)\\
        \bottomrule
    \end{tabular}
    }
    \captionsetup{font=small,labelfont=bf}
    \caption{ 
    This table is divided into three sets of rows with FP32 baselines at the top. 
    We provide two baselines for INT8 and INT4: standard QAT and stochastic QAT (nQAT).
    Both are informed by the analysis in \ref{sec:STEconfig}, with nQAT achieving better performance in some cases.
    Models trained with Degree-Quant (DQ) are always comparable to baselines, and usually substantially better, especially for INT4.
    \textbf{DQ is a stable method which requires little tuning to obtain excellent results across a variety of architectures and datasets.}
    }
    \label{tab:results}
\end{table*}

\subsection{Obtaining Quantization baselines}

Our FP32 results, which we obtain after extensive hyperparameter tuning, and those from the baselines are shown at the top of \cref{tab:results}.
We observed large gains on MNIST, CIFAR10 and, ZINC.

For our QAT-INT8 and QAT-INT4 baselines, we use the quantization configurations informed by our analysis in \cref{sec:STEconfig}.
For Citeseer we use the best resulting setup analysed for Cora, and for CIFAR-10 that from MNIST.
Then, the hyperparameters for each experiment were fine tuned individually, including noise rate $n\in[0.5, 0.95]$ for nQAT experiments. 
QAT-INT8 and QAT-INT4 results in \cref{tab:results} and QAT-INT4, with the exception of MNIST (an easy to classify dataset), corroborate our hypothesis that GIN layers are less resilient to quantization.
This was first observed in ~\cref{fig:indegree}.
In the case of ZINC, while all models results in noticeable degradation, GIN sees a more severe $16\%$ increase of regression loss compared to our FP32 baseline. For QAT W4A4 an accuracy drop of over $35\%$ and $47\%$ is observed for Cora and Citeseer respectively.
The stochasticity induced by nQAT helped in recovering some of the accuracy lost as a result of quantization for citation networks (both INT8 and INT4) but had little impact on other datasets and harmed performance in some cases.

\subsection{Comparisons of Degree-Quant with Existing Quantization Approaches}

Degree-Quant provides superior quantization for all GNN datasets and architectures. Our results with DQ are highlighted in gray in \cref{tab:results} and \cref{tab:resultsReddit}.
Citation networks trained with DQ for W8A8 manage to recover most of the accuracy lost as a result of QAT and outperform most of nQAT baselines.
In some instances DQ-W8A8 models outperform the reference FP32 baselines.
At 4-bits, DQ results in even larger gains compared to W4A4 baselines. 
We see DQ being more effective for GIN layers, outperforming INT4 baselines for Cora (+$24.9\%$), Citeseer (+$26.2\%$) and REDDIT-BINARY (+$23.0\%$) by large margins.
Models trained with DQ at W4A4 for graph classification and graph regression also exhibit large performance gains (of over $10\%$) in most cases.
For ZINC, all models achieve over $20\%$ lower regression loss.
Among the top performing models using DQ, ratios of $p_{\min}$ and $p_{\max}$ in $[0.0, 0.2]$ were the most common. \Cref{fig:redditRatios} in the appendix shows validation loss curves for GIN models trained using different DQ probabilities on the REDDIT-BINARY dataset.

\begin{table*}
\centering
\begin{minipage}{0.4\textwidth}
  \centering
        \tablefontsize
        \setlength{\tabcolsep}{4pt}
        \begin{tabular}{l c c}
            \toprule
            Quantization & Model & REDDIT-BIN (Acc.~\%) $\uparrow$ \\
            \midrule
            Ref. (FP32) & GIN & $92.2\pm2.3$ \\
            Ours (FP32) & GIN & $92.0\pm1.5$ \\
            \midrule
            QAT-W8A8 & GIN & $76.1\pm7.5$ \\
            nQAT-W8A8 & GIN & $77.5\pm3.4$ \\
            \rowcolor{Gray}
            DQ-W8A8 & GIN & $91.8\pm2.3$ (+$\bm{14.3}$) \\
            \midrule
            \rowcolor{white}
            QAT-W4A4 & GIN & $54.4\pm6.6$ \\
            nQAT-W4A4 & GIN & $58.0\pm6.3$ \\
            \rowcolor{Gray}
            DQ-W4A4 & GIN & $81.3\pm4.4$ (+$\bm{23.0}$) \\
            \bottomrule
        \end{tabular}
    \captionsetup{font=small,labelfont=bf}
    \captionof{table}{Results for DQ-INT8 GIN models perform nearly as well as at FP32. For INT4, DQ offers a significant increase in accuracy.
    }
    \label{tab:resultsReddit}
\end{minipage}\hfill%
\begin{minipage}{0.55\textwidth}
\centering
\tablefontsize
\setlength{\tabcolsep}{4pt}

\begin{tabular}{@{}cc|ccc|ccc@{}}
\toprule
\multirow{2}{*}{Device} & \multirow{2}{*}{Arch.} & \multicolumn{3}{c|}{Zinc (Batch=10K)} & \multicolumn{3}{c}{Reddit} \\
                        &                        & FP32 & W8A8 & Speedup     & FP32  & W8A8 & Speedup     \\ \midrule
\multirow{3}{*}{CPU}    & GCN                    & 181ms    & 42ms & 4.3$\times$ & 13.1s     & 3.1s    & 4.2$\times$ \\
                        & GAT                    & 190ms    & 50ms & 3.8$\times$ &  13.1s     & 2.8s    &  4.7$\times$ \\
                        & GIN                    & 182ms    & 43ms & 4.2$\times$ &     13.1s  & 3.1s & 4.2$\times$ \\ \midrule
\multirow{3}{*}{GPU}    & GCN                    & 39ms & 31ms & 1.3$\times$ & 191ms & 176ms & 1.1$\times$ \\
                        & GAT                    & 17ms & 15ms & 1.1$\times$ & OOM   & OOM  & -           \\
                        & GIN                    & 39ms & 31ms  & 1.3$\times$ & 191ms      & 176ms     & 1.1$\times$            \\ \bottomrule
\end{tabular}

\captionsetup{font=small,labelfont=bf}
    \captionof{table}{INT8 latency results run on a 22 core 2.1GHz Intel Xeon Gold 6152 and, on a GTX 1080Ti GPU. 
    Quantization provides large speedups on a variety of graphs for CPU and non-negligible speedups with unoptimized INT8 GPU kernels.
    }
    \label{tab:latency}
\end{minipage}
        
\end{table*}

\section{Discussion}



\textbf{Latency and Memory Implications}.
\label{sec:performance_implications}
In addition to offering significantly lower memory usage ($4\times$ with INT8), quantization can reduce latency---especially on CPUs.
We found that with INT8 arithmetic we could accelerate inference by up to $\mathbf{4.7}\times$.
We note that the latency benefit depends on the graph topology and feature dimension, therefore we ran benchmarks on a variety of graph datasets, including Reddit\footnote{The largest graph commonly benchmarked on in the GNN literature}, Zinc, Cora, Citeseer, and CIFAR-10; Zinc and Reddit results are shown in \cref{tab:latency}, with further results given in the appendix.
For a GCN layer with in- and out-dimension of 128, we get speed-ups of: $4.3\times$ on Reddit, $2.5\times$ on Zinc, $1.3\times$ on Cora, $1.3\times$ on Citeseer and, $2.1\times$ on CIFAR-10.
It is also worth emphasizing that quantized networks are necessary to efficiently use accelerators deployed in smartphones and smaller devices as they primarily accelerate integer arithmetic, and that CPUs remain a common choice for model serving on servers.
The decrease in latency on CPUs is due to improved cache performance for the sparse operations; GPUs, however, see less benefit due to their massively-parallel nature which relies on mechanisms other than caching to hide slow random memory accesses, which are unavoidable in this application.

\begin{figure}[h]
\centering
\begin{minipage}{.50\textwidth}
  \centering
  \includegraphics[width=0.9\linewidth]{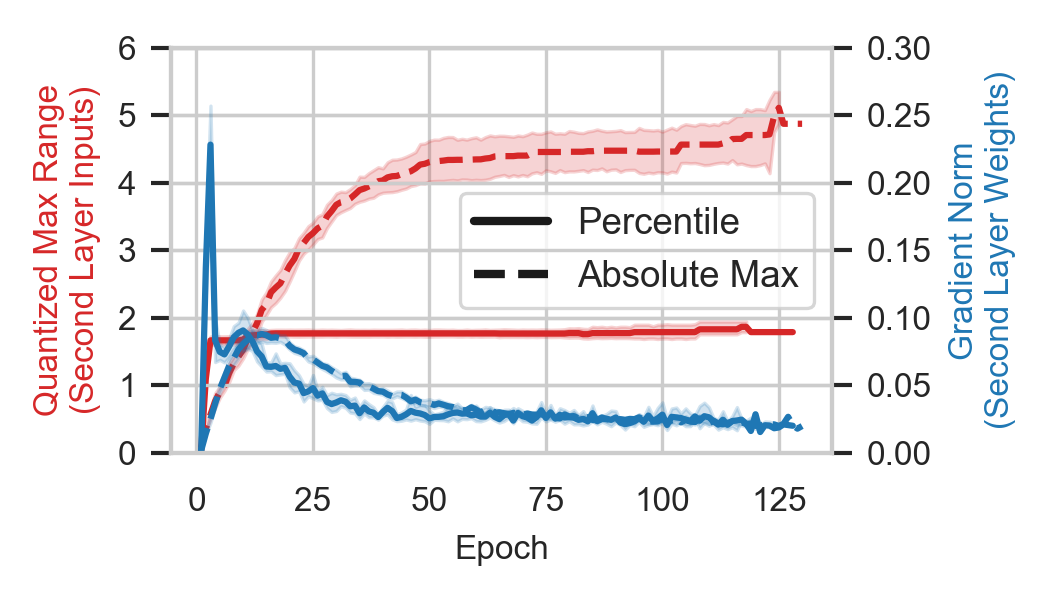}
  \captionsetup{font=small,labelfont=bf}
  \captionof{figure}{$q_{\max}$ with absolute min/max and percentile ranges, applied to INT8 GCN training on Cora.
  We observe that the percentile max is half that of the absolute, \emph{doubling} resolution for the majority of values.}
  \label{fig:norm_min_max}
\end{minipage}\hfill%
\begin{minipage}{.46\textwidth}
  \centering
  \includegraphics[width=0.94\linewidth]{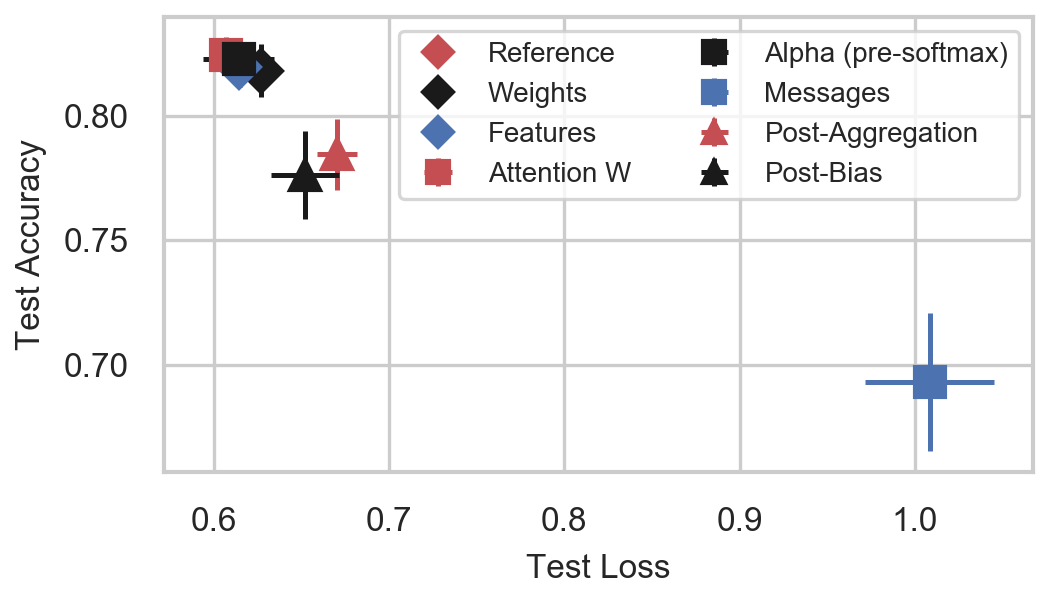}
  \captionsetup{font=small,labelfont=bf}
  \captionof{figure}{Analysis of how INT8 GAT performance degrades on Cora as individual elements are reduced to 4-bit precision \emph{without DQ}.
  For GAT the message elements are crucial to classification performance.}
  \label{fig:gat_deg}
\end{minipage}
\end{figure}

\textbf{Ablation Study: Benefits of Percentile Ranges}.
\Cref{fig:norm_min_max} shows the value of percentiles during training.
We see that when using absolute min/max the upper range grows to over double the range required for 99.9\% of values, effectively halving the resolution of the quantized values.
DQ is more stable, and we obtained strong results with an order of magnitude less tuning relative to the baselines. 

\textbf{Ablation Study: Source of Degradation at INT4}.
\label{sec:int4_ablate}
\Cref{fig:gat_deg} assesses how INT8 GAT (without DQ) degrades as single elements are converted to INT4, in order to understand the precipitous drop in accuracy in the INT4 baselines; further plots for GCN and GIN are included in the appendix.
We observe that most elements cause only modest performance losses relative to a full INT8 model.
DQ is most important to apply to elements which are constrained by \emph{numerical precision}, such as the aggregation and message elements in GAT.
Weight elements, however, are consistently unaffected.

\textbf{Ablation Study: Effect of Stochastic Element in Degree-Quant}.
We observe that the stochastic protective masking in DQ alone often achieves most of the performance gain over the QAT baseline; results are given in \cref{tab:stochDQ} in the appendix.
The benefit of the percentile-based quantization ranges is \emph{stability}, although it can yield some performance gains.
The full DQ method provides consistently good results on all architectures and datasets, without requiring an extensive analysis as in \ref{sec:STEconfig}.

\section{Conclusion}
This work has presented Degree-Quant, an architecture-agnostic and stable method for training quantized GNN models that can be accelerated using off-the-shelf hardware.
With 4-bit weights and activations we achieve $8\times$ compression while surpassing strong baselines by margins regularly exceeding 20\%.
At 8-bits, models trained with DQ perform on par or better than the baselines while achieving up to 4.7$\times$ lower latency than FP32 models.
Our work offers a comprehensive foundation for future work in this area and is a first step towards enabling GNNs to be deployed more widely, including to resource constrained devices such as smartphones.

\section*{Acknowledgements}
This work was supported by Samsung AI and by the UK’s Engineering and Physical Sciences Research Council (EPSRC) with grants EP/M50659X/1 and EP/S001530/1 (the MOA project) and the European Research Council via the REDIAL project.

\small
\bibliography{references}

\begin{thebibliography}{47}
\providecommand{\natexlab}[1]{#1}
\providecommand{\url}[1]{\texttt{#1}}
\expandafter\ifx\csname urlstyle\endcsname\relax
  \providecommand{\doi}[1]{doi: #1}\else
  \providecommand{\doi}{doi: \begingroup \urlstyle{rm}\Url}\fi

\bibitem[{Achanta} et~al.(2012){Achanta}, {Shaji}, {Smith}, {Lucchi}, {Fua},
  and {Süsstrunk}]{6205760}
R.~{Achanta}, A.~{Shaji}, K.~{Smith}, A.~{Lucchi}, P.~{Fua}, and
  S.~{Süsstrunk}.
\newblock Slic superpixels compared to state-of-the-art superpixel methods.
\newblock \emph{IEEE Transactions on Pattern Analysis and Machine
  Intelligence}, 34\penalty0 (11):\penalty0 2274--2282, 2012.

\bibitem[Alizadeh et~al.(2019)Alizadeh, Fernández-Marqués, Lane, and
  Gal]{alizadeh2018a}
Milad Alizadeh, Javier Fernández-Marqués, Nicholas~D. Lane, and Yarin Gal.
\newblock A empirical study of binary neural networks' optimisation.
\newblock In \emph{International Conference on Learning Representations}, 2019.
\newblock URL \url{https://openreview.net/forum?id=rJfUCoR5KX}.

\bibitem[Alizadeh et~al.(2020)Alizadeh, Behboodi, van Baalen, Louizos,
  Blankevoort, and Welling]{alizadeh2020gradient}
Milad Alizadeh, Arash Behboodi, Mart van Baalen, Christos Louizos, Tijmen
  Blankevoort, and Max Welling.
\newblock Gradient $\ell_1$ regularization for quantization robustness.
\newblock \emph{arXiv preprint arXiv:2002.07520}, 2020.

\bibitem[Allamanis et~al.(2018)Allamanis, Brockschmidt, and
  Khademi]{allamanis2018learning}
Miltiadis Allamanis, Marc Brockschmidt, and Mahmoud Khademi.
\newblock Learning to represent programs with graphs.
\newblock In \emph{6th International Conference on Learning Representations,
  {ICLR} 2018, Vancouver, BC, Canada, April 30 - May 3, 2018, Conference Track
  Proceedings}. OpenReview.net, 2018.
\newblock URL \url{https://openreview.net/forum?id=BJOFETxR-}.

\bibitem[Bengio et~al.(2013)Bengio, Léonard, and
  Courville]{bengio2013estimating}
Yoshua Bengio, Nicholas Léonard, and Aaron Courville.
\newblock Estimating or propagating gradients through stochastic neurons for
  conditional computation, 2013.

\bibitem[Blalock et~al.(2020)Blalock, Ortiz, Frankle, and
  Guttag]{blalock2020state}
Davis Blalock, Jose Javier~Gonzalez Ortiz, Jonathan Frankle, and John Guttag.
\newblock What is the state of neural network pruning?, 2020.

\bibitem[Carmichael et~al.(2018)Carmichael, Langroudi, Khazanov, Lillie,
  Gustafson, and Kudithipudi]{carmichael2018deep}
Zachariah Carmichael, Hamed~F. Langroudi, Char Khazanov, Jeffrey Lillie,
  John~L. Gustafson, and Dhireesha Kudithipudi.
\newblock Deep positron: A deep neural network using the posit number system,
  2018.

\bibitem[Corso et~al.(2020)Corso, Cavalleri, Beaini, Liò, and
  Veličković]{corso2020principal}
Gabriele Corso, Luca Cavalleri, Dominique Beaini, Pietro Liò, and Petar
  Veličković.
\newblock Principal neighbourhood aggregation for graph nets, 2020.

\bibitem[Courbariaux et~al.(2015)Courbariaux, Bengio, and
  David]{courbariaux2015binaryconnect}
Matthieu Courbariaux, Yoshua Bengio, and Jean-Pierre David.
\newblock Binaryconnect: Training deep neural networks with binary weights
  during propagations, 2015.

\bibitem[Dong et~al.(2017)Dong, Ni, Li, Chen, Zhu, and Su]{dong2017learning}
Yinpeng Dong, Renkun Ni, Jianguo Li, Yurong Chen, Jun Zhu, and Hang Su.
\newblock Learning accurate low-bit deep neural networks with stochastic
  quantization, 2017.

\bibitem[Duvenaud et~al.(2015)Duvenaud, Maclaurin, Iparraguirre, Bombarell,
  Hirzel, Aspuru-Guzik, and Adams]{duvenaud2015convolutional}
David~K Duvenaud, Dougal Maclaurin, Jorge Iparraguirre, Rafael Bombarell,
  Timothy Hirzel, Al{\'a}n Aspuru-Guzik, and Ryan~P Adams.
\newblock Convolutional networks on graphs for learning molecular fingerprints.
\newblock In \emph{Advances in neural information processing systems}, pp.\
  2224--2232, 2015.

\bibitem[Dwivedi et~al.(2020)Dwivedi, Joshi, Laurent, Bengio, and
  Bresson]{dwivedi2020benchmarking}
Vijay~Prakash Dwivedi, Chaitanya~K. Joshi, Thomas Laurent, Yoshua Bengio, and
  Xavier Bresson.
\newblock Benchmarking graph neural networks, 2020.

\bibitem[Esser et~al.(2020)Esser, McKinstry, Bablani, Appuswamy, and
  Modha]{Esser2020LEARNED}
Steven~K. Esser, Jeffrey~L. McKinstry, Deepika Bablani, Rathinakumar Appuswamy,
  and Dharmendra~S. Modha.
\newblock Learned step size quantization.
\newblock In \emph{International Conference on Learning Representations}, 2020.
\newblock URL \url{https://openreview.net/forum?id=rkgO66VKDS}.

\bibitem[Fan et~al.(2020)Fan, Stock, Graham, Grave, Gribonval, Jegou, and
  Joulin]{fan2020training}
Angela Fan, Pierre Stock, Benjamin Graham, Edouard Grave, Remi Gribonval, Herve
  Jegou, and Armand Joulin.
\newblock Training with quantization noise for extreme model compression, 2020.

\bibitem[Feng et~al.(2020)Feng, Wang, Li, Yang, Peng, and
  Ding]{feng2020sgquant}
Boyuan Feng, Yuke Wang, Xu~Li, Shu Yang, Xueqiao Peng, and Yufei Ding.
\newblock Sgquant: Squeezing the last bit on graph neural networks with
  specialized quantization, 2020.

\bibitem[Fey \& Lenssen(2019)Fey and Lenssen]{fey2019fast}
Matthias Fey and Jan~Eric Lenssen.
\newblock Fast graph representation learning with pytorch geometric, 2019.

\bibitem[Gilmer et~al.(2017)Gilmer, Schoenholz, Riley, Vinyals, and
  Dahl]{gilmer17neural}
Justin Gilmer, Samuel~S. Schoenholz, Patrick~F. Riley, Oriol Vinyals, and
  George~E. Dahl.
\newblock Neural message passing for quantum chemistry.
\newblock \emph{CoRR}, abs/1704.01212, 2017.
\newblock URL \url{http://arxiv.org/abs/1704.01212}.

\bibitem[Hamilton et~al.(2017)Hamilton, Ying, and
  Leskovec]{hamilton2017inductive}
William~L. Hamilton, Rex Ying, and Jure Leskovec.
\newblock Inductive representation learning on large graphs, 2017.

\bibitem[Han et~al.(2015)Han, Mao, and Dally]{han2015deep}
Song Han, Huizi Mao, and William~J. Dally.
\newblock Deep compression: Compressing deep neural networks with pruning,
  trained quantization and huffman coding, 2015.

\bibitem[Ioffe \& Szegedy(2015)Ioffe and Szegedy]{ioffe2015batch}
Sergey Ioffe and Christian Szegedy.
\newblock Batch normalization: Accelerating deep network training by reducing
  internal covariate shift, 2015.

\bibitem[Jacob et~al.(2017)Jacob, Kligys, Chen, Zhu, Tang, Howard, Adam, and
  Kalenichenko]{jacob2017quantization}
Benoit Jacob, Skirmantas Kligys, Bo~Chen, Menglong Zhu, Matthew Tang, Andrew
  Howard, Hartwig Adam, and Dmitry Kalenichenko.
\newblock Quantization and training of neural networks for efficient
  integer-arithmetic-only inference, 2017.

\bibitem[Jia et~al.(2020)Jia, Lin, Gao, Zaharia, and Aiken]{mlsys2020_83}
Zhihao Jia, Sina Lin, Mingyu Gao, Matei Zaharia, and Alex Aiken.
\newblock Improving the accuracy, scalability, and performance of graph neural
  networks with roc.
\newblock In \emph{Proceedings of Machine Learning and Systems 2020}, pp.\
  187--198. 2020.

\bibitem[Jin et~al.(2018)Jin, Barzilay, and Jaakkola]{jin2018junction}
Wengong Jin, Regina Barzilay, and Tommi Jaakkola.
\newblock Junction tree variational autoencoder for molecular graph generation,
  2018.

\bibitem[Kalamkar et~al.(2019)Kalamkar, Mudigere, Mellempudi, Das, Banerjee,
  Avancha, Vooturi, Jammalamadaka, Huang, Yuen, Yang, Park, Heinecke,
  Georganas, Srinivasan, Kundu, Smelyanskiy, Kaul, and
  Dubey]{kalamkar2019study}
Dhiraj Kalamkar, Dheevatsa Mudigere, Naveen Mellempudi, Dipankar Das, Kunal
  Banerjee, Sasikanth Avancha, Dharma~Teja Vooturi, Nataraj Jammalamadaka,
  Jianyu Huang, Hector Yuen, Jiyan Yang, Jongsoo Park, Alexander Heinecke,
  Evangelos Georganas, Sudarshan Srinivasan, Abhisek Kundu, Misha Smelyanskiy,
  Bharat Kaul, and Pradeep Dubey.
\newblock A study of bfloat16 for deep learning training, 2019.

\bibitem[Kipf \& Welling(2017)Kipf and Welling]{kipf2017semi}
Thomas~N. Kipf and Max Welling.
\newblock Semi-supervised classification with graph convolutional networks.
\newblock In \emph{5th International Conference on Learning Representations,
  {ICLR} 2017, Toulon, France, April 24-26, 2017, Conference Track
  Proceedings}. OpenReview.net, 2017.
\newblock URL \url{https://openreview.net/forum?id=SJU4ayYgl}.

\bibitem[Krishnamoorthi(2018)]{krishnamoorthi2018quantizing}
Raghuraman Krishnamoorthi.
\newblock Quantizing deep convolutional networks for efficient inference: A
  whitepaper, 2018.

\bibitem[Li et~al.(2020)Li, Jamieson, Rostamizadeh, Gonina, Ben-tzur, Hardt,
  Recht, and Talwalkar]{async_hyperband}
Liam Li, Kevin Jamieson, Afshin Rostamizadeh, Ekaterina Gonina, Jonathan
  Ben-tzur, Moritz Hardt, Benjamin Recht, and Ameet Talwalkar.
\newblock A system for massively parallel hyperparameter tuning.
\newblock In \emph{Proceedings of Machine Learning and Systems 2020}, pp.\
  230--246. 2020.

\bibitem[Louizos et~al.(2017)Louizos, Ullrich, and
  Welling]{louizos2017bayesian}
Christos Louizos, Karen Ullrich, and Max Welling.
\newblock Bayesian compression for deep learning, 2017.

\bibitem[Micikevicius et~al.(2017)Micikevicius, Narang, Alben, Diamos, Elsen,
  Garcia, Ginsburg, Houston, Kuchaiev, Venkatesh, and
  Wu]{micikevicius2017mixed}
Paulius Micikevicius, Sharan Narang, Jonah Alben, Gregory Diamos, Erich Elsen,
  David Garcia, Boris Ginsburg, Michael Houston, Oleksii Kuchaiev, Ganesh
  Venkatesh, and Hao Wu.
\newblock Mixed precision training, 2017.

\bibitem[Mukkara et~al.(2018)Mukkara, Beckmann, Abeydeera, Ma, and
  Sanchez]{10.1109/MICRO.2018.00010}
Anurag Mukkara, Nathan Beckmann, Maleen Abeydeera, Xiaosong Ma, and Daniel
  Sanchez.
\newblock Exploiting locality in graph analytics through hardware-accelerated
  traversal scheduling.
\newblock In \emph{Proceedings of the 51st Annual IEEE/ACM International
  Symposium on Microarchitecture}, MICRO-51, pp.\  1–14. IEEE Press, 2018.
\newblock ISBN 9781538662403.
\newblock \doi{10.1109/MICRO.2018.00010}.
\newblock URL \url{https://doi.org/10.1109/MICRO.2018.00010}.

\bibitem[Prato et~al.(2019)Prato, Charlaix, and Rezagholizadeh]{prato2019fully}
Gabriele Prato, Ella Charlaix, and Mehdi Rezagholizadeh.
\newblock Fully quantized transformer for machine translation, 2019.

\bibitem[Rong et~al.(2020)Rong, Huang, Xu, and Huang]{rong2020dropedge}
Yu~Rong, Wenbing Huang, Tingyang Xu, and Junzhou Huang.
\newblock Dropedge: Towards deep graph convolutional networks on node
  classification.
\newblock In \emph{International Conference on Learning Representations}, 2020.
\newblock URL \url{https://openreview.net/forum?id=Hkx1qkrKPr}.

\bibitem[Sarlin et~al.(2019)Sarlin, DeTone, Malisiewicz, and
  Rabinovich]{sarlin2019superglue}
Paul-Edouard Sarlin, Daniel DeTone, Tomasz Malisiewicz, and Andrew Rabinovich.
\newblock Superglue: Learning feature matching with graph neural networks.
\newblock \emph{arXiv preprint arXiv:1911.11763}, 2019.

\bibitem[Sheng et~al.(2018)Sheng, Feng, Zhuo, Zhang, Shen, and
  Aleksic]{Sheng_2018}
Tao Sheng, Chen Feng, Shaojie Zhuo, Xiaopeng Zhang, Liang Shen, and Mickey
  Aleksic.
\newblock A quantization-friendly separable convolution for mobilenets.
\newblock \emph{2018 1st Workshop on Energy Efficient Machine Learning and
  Cognitive Computing for Embedded Applications (EMC2)}, Mar 2018.
\newblock \doi{10.1109/emc2.2018.00011}.
\newblock URL \url{http://dx.doi.org/10.1109/emc2.2018.00011}.

\bibitem[Shkolnik et~al.(2020)Shkolnik, Chmiel, Banner, Shomron, Nahshan,
  Bronstein, and Weiser]{shkolnik2020robust}
Moran Shkolnik, Brian Chmiel, Ron Banner, Gil Shomron, Yuri Nahshan, Alex
  Bronstein, and Uri Weiser.
\newblock Robust quantization: One model to rule them all.
\newblock \emph{arXiv preprint arXiv:2002.07686}, 2020.

\bibitem[Srivastava et~al.(2014)Srivastava, Hinton, Krizhevsky, Sutskever, and
  Salakhutdinov]{srivastava2014dropout}
Nitish Srivastava, Geoffrey Hinton, Alex Krizhevsky, Ilya Sutskever, and Ruslan
  Salakhutdinov.
\newblock Dropout: a simple way to prevent neural networks from overfitting.
\newblock \emph{The journal of machine learning research}, 15\penalty0
  (1):\penalty0 1929--1958, 2014.

\bibitem[van~den Berg et~al.(2017)van~den Berg, Kipf, and
  Welling]{berg2017graph}
Rianne van~den Berg, Thomas~N. Kipf, and Max Welling.
\newblock Graph convolutional matrix completion, 2017.

\bibitem[Velickovic et~al.(2018)Velickovic, Cucurull, Casanova, Romero,
  Li{\`{o}}, and Bengio]{velickovic2018graph}
Petar Velickovic, Guillem Cucurull, Arantxa Casanova, Adriana Romero, Pietro
  Li{\`{o}}, and Yoshua Bengio.
\newblock Graph attention networks.
\newblock In \emph{6th International Conference on Learning Representations,
  {ICLR} 2018, Vancouver, BC, Canada, April 30 - May 3, 2018, Conference Track
  Proceedings}. OpenReview.net, 2018.
\newblock URL \url{https://openreview.net/forum?id=rJXMpikCZ}.

\bibitem[Wang et~al.(2020)Wang, Lian, Zhang, Qin, He, Lin, and
  Lin]{wang2020binarized}
Hanchen Wang, Defu Lian, Ying Zhang, Lu~Qin, Xiangjian He, Yiguang Lin, and
  Xuemin Lin.
\newblock Binarized graph neural network, 2020.

\bibitem[Wang et~al.(2018)Wang, Liu, Lin, Lin, and Han]{wang2018haq}
Kuan Wang, Zhijian Liu, Yujun Lin, Ji~Lin, and Song Han.
\newblock Haq: Hardware-aware automated quantization with mixed precision,
  2018.

\bibitem[Wu et~al.(2020)Wu, Judd, Zhang, Isaev, and
  Micikevicius]{wu2020integer}
Hao Wu, Patrick Judd, Xiaojie Zhang, Mikhail Isaev, and Paulius Micikevicius.
\newblock Integer quantization for deep learning inference: Principles and
  empirical evaluation, 2020.

\bibitem[Wu et~al.(2017)Wu, Ramsundar, Feinberg, Gomes, Geniesse, Pappu,
  Leswing, and Pande]{wu2017moleculenet}
Zhenqin Wu, Bharath Ramsundar, Evan~N. Feinberg, Joseph Gomes, Caleb Geniesse,
  Aneesh~S. Pappu, Karl Leswing, and Vijay Pande.
\newblock Moleculenet: A benchmark for molecular machine learning, 2017.

\bibitem[Xu et~al.(2019)Xu, Hu, Leskovec, and Jegelka]{xu19how}
Keyulu Xu, Weihua Hu, Jure Leskovec, and Stefanie Jegelka.
\newblock How powerful are graph neural networks?
\newblock In \emph{7th International Conference on Learning Representations,
  {ICLR} 2019, New Orleans, LA, USA, May 6-9, 2019}. OpenReview.net, 2019.
\newblock URL \url{https://openreview.net/forum?id=ryGs6iA5Km}.

\bibitem[Yan et~al.(2020)Yan, Wang, Guo, and Lou]{10.1145/3394486.3403236}
Bencheng Yan, Chaokun Wang, Gaoyang Guo, and Yunkai Lou.
\newblock Tinygnn: Learning efficient graph neural networks.
\newblock In \emph{Proceedings of the 26th ACM SIGKDD International Conference
  on Knowledge Discovery and Data Mining}, KDD '20, pp.\  1848–1856, New
  York, NY, USA, 2020. Association for Computing Machinery.
\newblock ISBN 9781450379984.
\newblock \doi{10.1145/3394486.3403236}.
\newblock URL \url{https://doi.org/10.1145/3394486.3403236}.

\bibitem[Zafrir et~al.(2019)Zafrir, Boudoukh, Izsak, and
  Wasserblat]{zafrir2019q8bert}
Ofir Zafrir, Guy Boudoukh, Peter Izsak, and Moshe Wasserblat.
\newblock Q8bert: Quantized 8bit bert, 2019.

\bibitem[Zeng \& Prasanna(2020)Zeng and Prasanna]{Zeng_2020}
Hanqing Zeng and Viktor Prasanna.
\newblock Graphact.
\newblock \emph{The 2020 ACM/SIGDA International Symposium on
  Field-Programmable Gate Arrays}, Feb 2020.
\newblock \doi{10.1145/3373087.3375312}.
\newblock URL \url{http://dx.doi.org/10.1145/3373087.3375312}.

\bibitem[Zeng et~al.(2020)Zeng, Zhou, Srivastava, Kannan, and
  Prasanna]{graphsaint}
Hanqing Zeng, Hongkuan Zhou, Ajitesh Srivastava, Rajgopal Kannan, and Viktor
  Prasanna.
\newblock Graphsaint: Graph sampling based inductive learning method, 2020.

\end{thebibliography}
\bibliographystyle{iclr2021_conference}

\cleardoublepage
\normalsize
\appendix
\section{Appendix}
\label{appendix}

Readers seeking advice on implementation will find \cref{sec:impl_advice} especially useful.
We provide significant advice surrounding best practices on quantization for GNNs, along with techniques which we believe can boost our methods beyond the performance described in this paper, but for which we did not have time to fully evaluate.

\subsection{Experimental Setup}
\label{app:experimentalSetup}

As baselines we use the architectures and results reported by \citet{fey2019fast} for citation networks, \citet{dwivedi2020benchmarking} for MNIST, CIFAR-10 and ZINC and, \citet{xu19how} for REDDIT-BINARY.
We re-implemented the architectures and datasets used in these publications and replicated the results reported at FP32.
Models using GIN layers learn parameter \large{$\epsilon$}. \normalsize 
These models are often referred to as GIN-\large{$\epsilon$}. \normalsize 
The high-level description of these architectures is shown in \cref{tab:archDescription}.
The number of parameters for each architecture-dataset in this work are shown in \cref{tab:archSizes}.

Our infrastructure was implemented using PyTorch Geometric (PyG)~\citep{fey2019fast}.
We generate candidate hyperparameters using random search, and prune trials using the asynchronous hyperband algorithm~\citep{async_hyperband}.
Hyperparameters searched over were learning rate, weight decay, and dropout~\citep{srivastava2014dropout} and drop-edge~\citep{rong2020dropedge} probabilities. The search ranges were initialized centered at the values used in the reference implementations of the baselines.
Degree-Quant requires searching for two additional hyperparameters, $p_{\min}$ and $p_{\max}$, these were tuned in a grid-search fashion.
We report our results using the hyperparameters which achieved the best validation loss over 100 runs on the Cora and Citeseer datasets, 10 runs for MNIST, CIFAR-10 and ZINC, and 10-fold cross-validation for REDDIT-BINARY.

We generally used fewer hyperparameter runs for our DQ runs than we did for baselines---even ignoring the searches over the various STE configs.
As our method is more stable, finding a reasonable set of parameters was easier than before.
As is usual with quantization experiments, we found it useful to decrease the learning rate relative to the FP32 baseline.

Our experiments ran on several machines in our SLURM cluster using Intel CPUs and NVIDIA GPUs.
Each machine was running Ubuntu 18.04.
The GPU models in our cluster were: V100, RTX 2080Ti and GTX 1080Ti.

\begin{table*}[ht]
    \tablefontsize
    \centering
    \setlength{\tabcolsep}{5pt}
    \begin{tabular}{l | c c c c c | c c c c c | c c c c c | c c c c c}
        \toprule
        Model & \multicolumn{5}{c}{\# Layers} & \multicolumn{5}{c}{\# Hidden Units} & \multicolumn{5}{c}{Residual} & \multicolumn{5}{c}{Output MLP} \\
        Arch. & Cit & M & C & Z & R & Cit & M & C & Z & R & Cit & M & C & Z & R & Cit & M & C & Z & R  \\
        \midrule
        GCN & 2 & 4 & 4 & 4 & -     & 16 & 146 & 146 & 145 & -      & $\times$ & \checkmark & \checkmark & \checkmark & -       & $\times$ & \checkmark & \checkmark & \checkmark & - \\
        GAT & 2 & 4 & 4 & 4 & -     & 8 & 19 & 19 & 18 & -          & $\times$ & \checkmark & \checkmark & \checkmark & -       & $\times$ & \checkmark & \checkmark & \checkmark & - \\
        GIN & 2 & 4 & 4 & 4 & 5     & 16 & 110 & 110 & 110 & 64     & $\times$ & \checkmark & \checkmark & \checkmark & $\times$    & $\times$ & \checkmark & \checkmark & \checkmark & \checkmark \\
        \bottomrule
    \end{tabular}
    \captionsetup{font=small,labelfont=bf}
    \caption{High level description of the architectures evaluated for citation networks (Cit), MNIST (M), CIFAR-10 (C), ZINC (Z) and REDDIT-BINARY (R). We relied on Adam optimizer for all experiments. For all batched experiments, we used 128 batch-sizes. All GAT models used 8 attention heads. All GIN architectures used 2-layer MLPs, except those for citation networks which used a single linear layer.}
    \label{tab:archDescription}
\end{table*}

\begin{table*}[ht]
    \small
    \centering
    \begin{tabular}{l c c c c c c}
        \toprule
        Model & \multicolumn{2}{c}{Node Classification} & \multicolumn{3}{c}{Graph Classification} & \multicolumn{1}{c}{Graph Regression} \\
        
        Arch. & Cora & Citeseer & MNIST & CIFAR-10 & REDDIT-BIN  & ZINC  \\
        \midrule
        GCN & 23063 & 59366 & 103889 & 104181 & - & 105454 \\
        GAT & 92373 & 237586 & 113706 & 114010 & - & 105044 \\
        GIN & 23216 & 59536 & 104554 & 104774 & 42503 & 102088 \\
        \bottomrule
    \end{tabular}
    \captionsetup{font=small,labelfont=bf}
    \caption{Number of parameters for each of the evaluated architectures}
    \label{tab:archSizes}
\end{table*}

For QAT experiments, all elements of each network are quantized: inputs to each layer, the weights, the messages sent between nodes, the inputs to aggregation stage and its outputs and, the outputs of the update stage (which are the outputs of the GNN layer before activation). In this way, all intermediate tensors in GNNs are quantized with the exception of the attention mechanism in GAT; we do not quantize after the softmax calculation, due to the numerical precision required at this stage. With the exception of Cora and Citeseer, the models evaluated in this work make use of Batch Normalization~\citep{ioffe2015batch}. For deployments of quantized models, Batch Normalization layers are often \textit{folded} with the weights~\citep{krishnamoorthi2018quantizing}. This is to ensure the input to the next layer is within the expected $[q_{\min},q_{\max}]$ ranges. In this work, for both QAT baselines and QAT+DQ, we left BN layers unfolded but ensure the inputs and outputs were quantized to the appropriate number of bits (i.e. INT8 or INT4) before getting multiplied with the layer weights. We leave as future work proposing a BN folding mechanism applicable for GNNs and studying its impact for deployments of quantized GNNs.

The GIN models evaluated on REDDIT-BINARY used QAT for all layers with the exception of the input layer of the MLP in the first GIN layer. This compromise was needed to overcome the severe degradation introduced by quantization when operating on nodes with a single scalar as feature. 

\subsection{Datasets}
We show in \Cref{tab:statsDatasets} the statistics for each dataset either used or referred to in this work. 
For Cora and Citeseer datasets, nodes correspond to documents and edges to citations between these.
Node features are a bag-of-words representation of the document. 
The task is to classify each node in the graph (i.e. each document) correctly.
The MNIST and CIFAR-10 datasets (commonly used for image classification) are transformed using SLIC~\citep{6205760} into graphs where each node represents a cluster of perceptually similar pixels or superpixels.
The task is to classify each image using their superpixels graph representation.
The ZINC dataset contains graphs representing molecules, were each node is an atom. 
The task is to regress a molecular property (constrained solubility~\citep{jin2018junction}) given the graph representation of the molecule.
Nodes in graphs of the REDDIT-BINARY dataset represent users of a Reddit thread with edges drawn between a pair of nodes if these interacted. 
This dataset contains graphs of two types of communities: question-answer threads and discussion threads.
The task is to determine if a given graph is from a question-answer thread or a discussion thread.

We use standard splits for MNIST, CIFAR-10 and ZINC.
For citation datasets (Cora and Citeseer), we use the splits used by \citet{kipf2017semi}.
For REDDIT-BINARY we use 10-fold cross validation.


\begin{table*}[h]
    \tablefontsize
    \centering
    \begin{tabular}{cccccc}
    \toprule
    Dataset       & Graphs & Nodes        & Edges       & Features & Labels \\ \midrule
    Cora          & 1      & 2,708        & 5,278       & 1,433    & 7      \\
    Citeseer      & 1      & 3,327        & 4,552       & 3,703    & 6      \\
    Pubmed        & 1      & 19,717       & 44,324      & 500      & 3      \\
    MNIST         & 70K    & 40-75        & 564.53 (avg)& 3        & 10     \\
    CIFAR10       & 60K    & 85-150       & 941.07 (avg)& 5        & 10     \\
    ZINC          & 12K    & 9-37         & 49.83 (avg) & 28       & 1      \\
    REDDIT-BINARY & 2K     & 429.63 (avg) & 497.75 (avg)& 1        & 2      \\
    Reddit        & 1      & 232,965      & 114,848,857 & 602      & 41     \\
    Amazon        & 1      & 9,430,088    & 231,594,310 & 300      & 24     \\ \bottomrule
    \end{tabular}
    \captionsetup{font=small,labelfont=bf}
    \caption{Statistics for each dataset used in the paper.
    Some datasets are only referred to in \cref{fig:OPs}}
    \label{tab:statsDatasets}
\end{table*}

\subsection{Quantization Implementations}
\label{app:STEconfig}


In \cref{sec:STEconfig} we analyse different readily available quantization implementations and how they impact in QAT results. First, vanilla STE, which is the reference STE~\citep{bengio2013estimating} that lets the gradients pass unchanged; and gradient clipping (GC), which clips the gradients based on the maximum representable value for a given quantization level. Or in other words, GC limits gradients if the tensor's magnitudes are outside the [$q_{\min}$, $q_{\max}$] range.

\begin{equation}
 x_{\min} = 
  \begin{cases} 
   \min(X) & \text{if step} = 0 \\
   \min(x_{\min}, X)       & \text{otherwise}
  \end{cases}
  \label{eq:min/max}
\end{equation}

\begin{equation}
 x_{\min} = 
  \begin{cases} 
   \min(X) & \text{if step } = 0 \\
   (1-c)x_{\min} +c\min(X)       & \text{otherwise} 
  \end{cases}
  \label{eq:mommin/max}
\end{equation}

The quantization modules keep track of the input tensor’s min and max values, $x_{\min}$ and $x_{\max}$, which are then used to compute $q_{\min}$, $q_{\max}$, \textit{zero-point} and \textit{scale} parameters. For both vanilla STE and GC, we study two popular ways of keeping track of these statistics: \textit{min/max}, which tracks the min/max tensor values observed over the course of training; and \textit{momentum}, which computes the moving averages of those statistic during training. The update rules for $x_{\min}$ for STE \textit{min/max} and STE \textit{momentum} are presented in \cref{eq:min/max} and \cref{eq:mommin/max} respectively, where $X$ is the tensor to be quantized and $c$ is the momentum hyperparameter, which in all our experiments is set to its default $0.01$. Equivalent rules apply when updating $x_{\max}$ (omitted).

For stochastic QAT we followed the implementation described in \citet{fan2020training}, where at each training step a binary mask sampled from a Bernoulli distribution is used to specify which elements of the weight tensor will be quantized and which will be left at full precision. We experimented with block sizes larger than one (i.e. a single scalar) but often resulted in a sever drop in performance. All the reported results use block size of one.

\subsection{Degree-Quant and Graph Level Summarization}
The percentile operation in our quantization scheme remains important for summarizing the graph when doing graph-level tasks, such as graph regression (Zinc) or graph classification (MNIST, CIFAR-10 and REDDIT-BINARY).
Since the number of nodes in each input graph is not constant, this can cause the summarized representation produced from the final graph layer to have a more tailed distribution than would be seen with other types of architectures (e.g. CNN).
Adding the percentile operation reduces the impact of these extreme tails in the fully connected graph-summarization layers, thereby increasing overall performance.
The arguments regarding weight update accuracy also still apply, as the $\frac{\partial \mathcal{L}}{\partial \mathbf{h}^{(i)}_{l+1}}$ term in the equations for the GCN and GIN should be more accurate compared to when the activations are always quantized before the summarization.
This phenomenon is also noted by \citet{fan2020training}.

\subsection{Implementation Advice}\label{sec:impl_advice}
We provide details that will be useful for others working in the area, including suggestions that should boost the performance of our results and accelerate training.
We release code on GitHub; this code is a clean implementation of the paper, suitable for users in downstream works.

\subsubsection{Quantization Setup}
As our work studies the pitfalls of quantization for GNNs, we were more aggressive in our implementation than is absolutely necessary: everything (where reasonably possible) in our networks is quantized.
In practice, this leaves low-hanging fruit for improvements in accuracy:

\begin{itemize}
    \item Not quantizing the final layer (as is common practice for CNNs and Transformers) helps with accuracy, especially at INT4.
    A similar practice at the first layer will also be useful.
    \item Using higher precision for the ``summarization'' stages of the model, which contributes little towards the runtime in most cases.
    \item Taking advantage of mixed precision: since the benefits of quantization are primarily in the message passing phase (discussed below), one technique to boost accuracy is to only make the messages low precision.
\end{itemize}

We advise choosing a more realistic (less aggressive) convention than used in this work.
The first two items would be appropriate.

\subsubsection{Relative Value of Percentiles Compared to Protective Masking}
There are two components to our proposed technique: stochastic, topology-aware, masking and percentile-based range observers for quantizers.
We believe that percentiles provide more immediate value, especially at INT4.
We find that they are useful purely from the perspective of stabilizing the optimization and reducing the sensitivity to hyperparameters.

However, adding the masking does improve performance further.
This is evident from \cref{tab:stochDQ}.
In fact, performance may be degraded slightly when percentiles are also applied: this can be observed by comparing \cref{tab:stochDQ} to the main results in the paper, \cref{tab:results}.

\subsubsection{Percentiles}
The key downside with applying percentiles for range observers is that the operation can take significant time.
Training with DQ is slower than before---however, since there is less sensitivity to hyperparameters, fewer runs end up being needed.
We are confident that an effective way to speed up this operation is to use sampling.
We expect 10\% of the data should be adequate, however we believe that even 1\% of the data may be sufficient (dataset and model dependent).
However, we have not evaluated this setup in the paper; it is provided in the code release for experimentation.

\subsubsection{Improving on Percentiles}
We believe that it is possible to significantly boost the performance of GNN quantization by employing a learned step size approach.
Although we used percentiles in this paper to illustrate the range-precision trade-off for GNNs, we expect that \textit{learning} the ranges will lead to better results.
This approach, pioneered by works such as \citet{Esser2020LEARNED}, has been highly effective on CNNs even down to 2 bit quantization.

Another approach would be to use \textit{robust quantization}: the ideas in these works are to reduce the impact of changing quantization ranges i.e.~making the architecture more robust to quantization.
Works in this area include \citet{alizadeh2020gradient} and \citet{shkolnik2020robust}.

\subsubsection{Improving Latency}
The slowest step of GNN inference is typically the sparse operations.
It is therefore best to minimize the sizes of the messages between nodes i.e.~quantize the message phase most aggressively.
This makes the biggest impact on CPUs which are dependent on caches to obtain good performance.

We evaluated our code on CPU using Numpy and Scipy routines.
For the GPU, we used implementations from PyTorch and PyTorch Geometric and lightly modified them to support INT8 where necessary.
These results, while useful for illustrating the benefits of quantization, are by no means optimal: we did not devote significant time to improving latency.
We believe better results can be obtained by taking advantage of techniques such as cache blocking or kernel fusion.

\subsubsection{Pitfalls}
Training these models can be highly unstable: some experiments in the paper had standard deviations as large as 18\%.
We observed this to affect citation network experiments to the extent that they would not converge on GPUs: all these experiments had to be run on CPUs.

\subsection{Degradation Studies}
\begin{figure}[h]
\centering
\begin{minipage}{.48\textwidth}
  \centering
  \includegraphics[width=\linewidth]{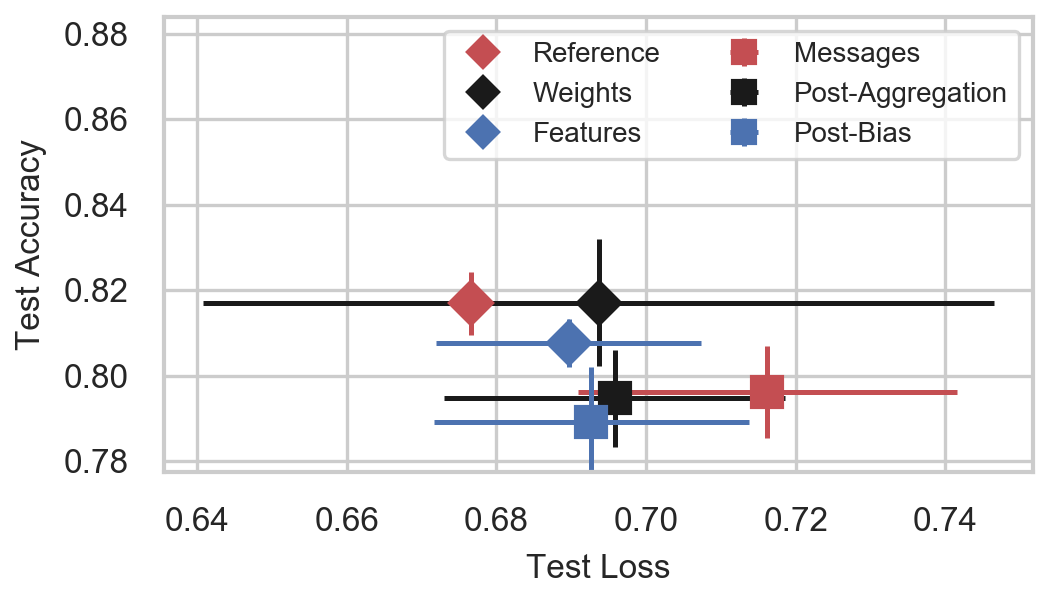}
  \captionsetup{font=small,labelfont=bf}
  \captionof{figure}{Degradation of INT8 GCN on Cora as individual elements are converted to INT4 \emph{without Degree-Quant}.}
  \label{fig:gcn_deg}
\end{minipage}\hfill%
\begin{minipage}{.48\textwidth}
  \centering
  \includegraphics[width=\linewidth]{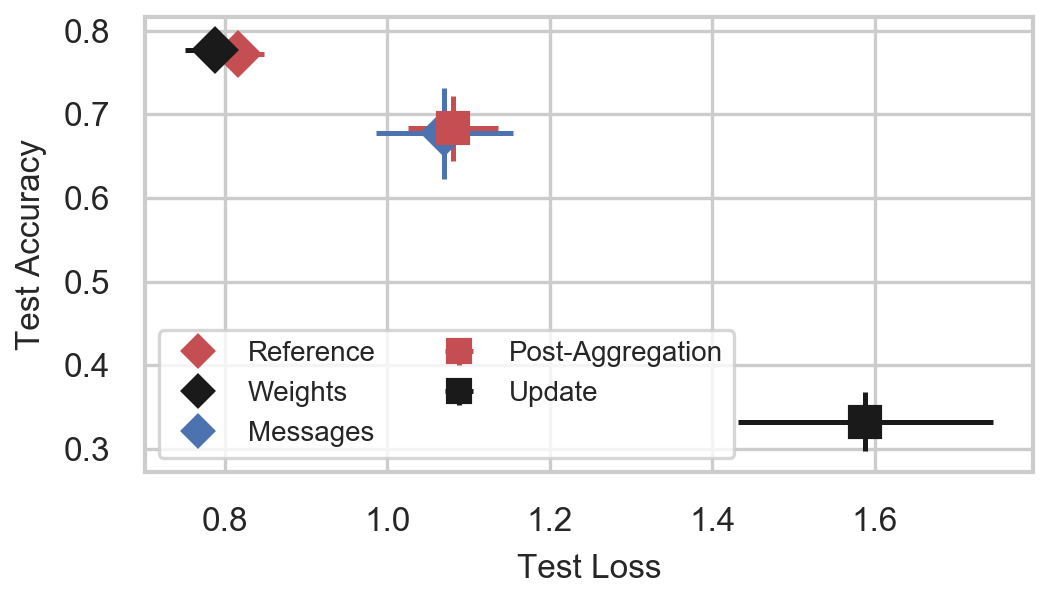}
  \captionsetup{font=small,labelfont=bf}
  \captionof{figure}{Degradation of INT8 GIN on Cora as individual elements are converted to INT4 \emph{without Degree-Quant}.}
  \label{fig:gin_deg}
\end{minipage}
\end{figure}

Figures \ref{fig:gcn_deg} and \ref{fig:gin_deg} show the results of the ablation study conducted in \cref{sec:int4_ablate} for GCN and GIN.
We observe that GCN is more tolerant to INT4 quantization than other architectures.
GIN, however, requires accurate representations after the update stage, and heavily suffers from further quantization like GAT.
The idea of performing different stages of inference at different precisions has been proposed, although it is uncommon~\citep{wang2018haq}.

\begin{figure}[h]
    \centering
    \includegraphics[width=0.9\textwidth]{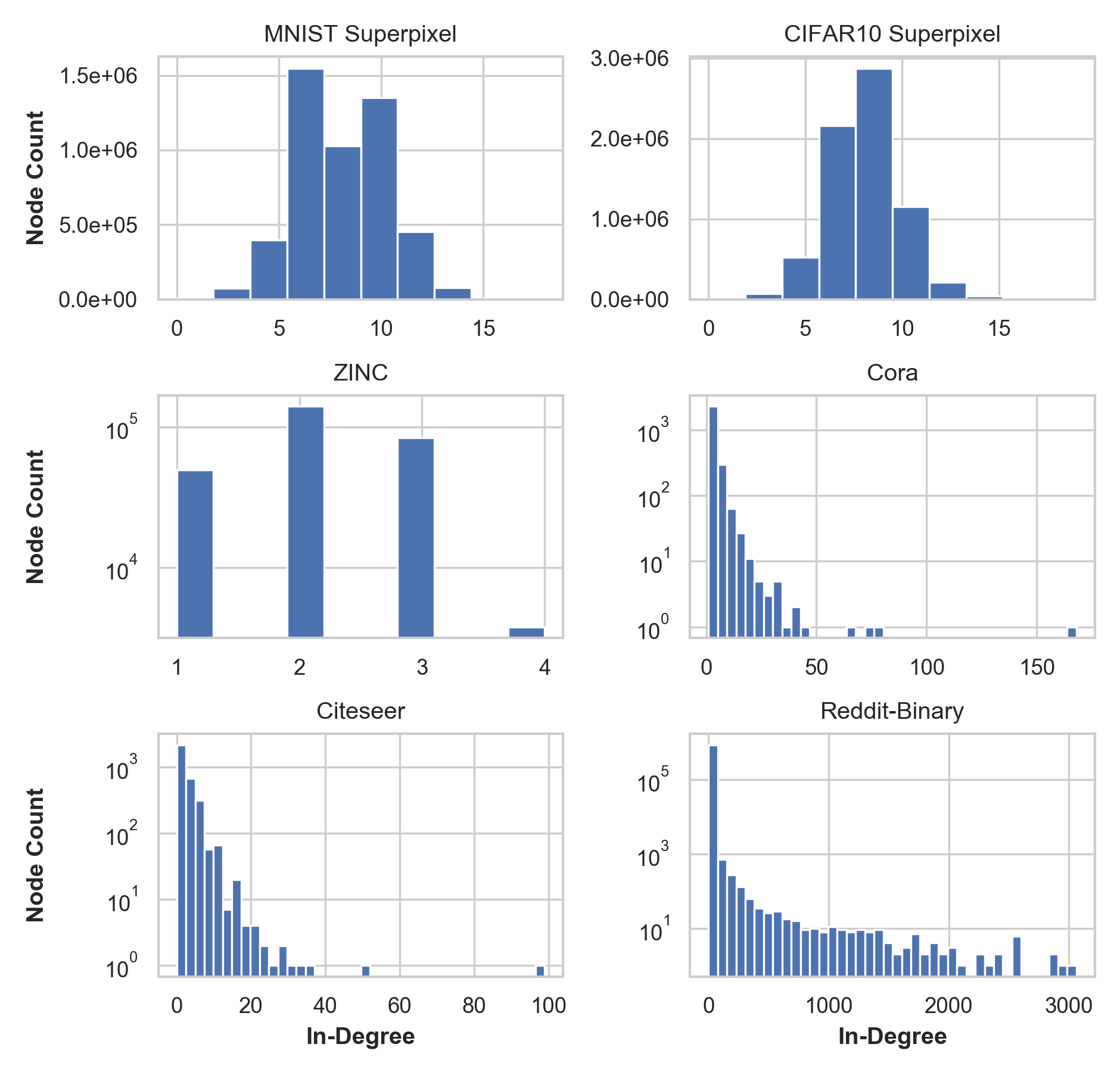}
    \captionsetup{font=small,labelfont=bf}
    \caption{In-degree distribution for each of the six datasets assessed.
    Note that a log $y$-axis is used for all datasets except for MNIST and CIFAR-10.}
    \label{fig:indegree_dist}
\end{figure}

\begin{figure}
\centering
    \begin{minipage}[c]{.6\textwidth}
    \centering
    \vspace{0pt}
    \includegraphics[width=\linewidth]{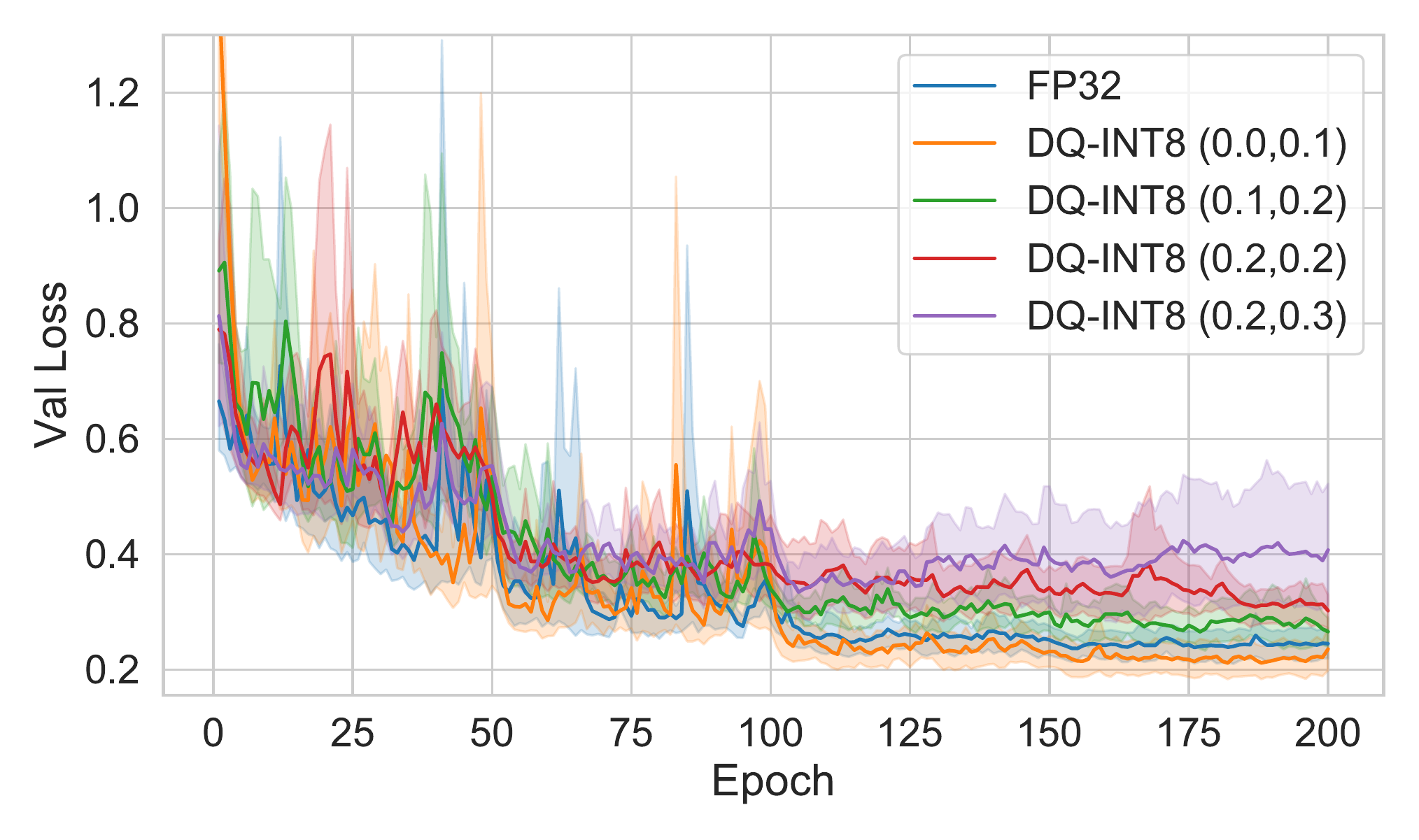}
    \captionsetup{font=small,labelfont=bf}
    \caption{Validation loss curves for GIN models evaluated on REDDIT-BINARY. Results averaged across 10-fold cross-validation. We show four DQ-INT8 experiments each with a different values for ($p_{\min}$,$p_{\max}$) and our FP32 baseline.}
    \label{fig:redditRatios}
    \end{minipage}
    \hfill
\begin{minipage}[c]{.38\textwidth}
    \vspace{10pt}
    \centering
    \tablefontsize
    \begin{tabular}{l c c}
            \toprule
            Quantization & Model & REDDIT-BIN $\uparrow$ \\
            \midrule
            Ref. (FP32) & GIN & $92.2\pm2.3$ \\[1mm]
            Ours (FP32) & GIN & $92.0\pm1.5$ \\[1mm]
            \midrule
            DQ-INT8 (0.0, 0.1) & GIN & $91.8\pm2.3$ \\[1mm]
            DQ-INT8 (0.1, 0.2) & GIN & $90.1\pm2.5$ \\[1mm]
            DQ-INT8 (0.2, 0.2) & GIN & $89.0\pm3.0$ \\[1mm]
            DQ-INT8 (0.2, 0.3) & GIN & $88.1\pm3.0$ \\
            \bottomrule
        \end{tabular}
    \label{tab:redditRatios}   
    \captionsetup{font=small,labelfont=bf}
    \captionof{table}{Final test accuracies for FP32 and DQ-INT8 models whose validation loss curves are shown in~\cref{fig:redditRatios}}
\end{minipage}
\end{figure}

\begin{table*}[ht]
    \tablefontsize
    \centering
    \begin{tabular}{l c c c c}
        \toprule
        Quantization & Model & \multicolumn{2}{c}{Node Classification} & \multicolumn{1}{c}{Graph Regression} \\
        
        Scheme & Arch. & Cora $\uparrow$ & Citeseer $\uparrow$ & ZINC $\downarrow$ \\
        \midrule
        \multirow{3}{*}{QAT-INT8 + DQ Masking} & GCN & $81.1\pm0.6$ & $71.0\pm0.7$ & $0.468\pm0.014$ \\
         & GAT & $82.1\pm0.1$ & $71.4\pm0.8$ & $0.462\pm0.005$ \\
         & GIN & $78.9\pm1.2$ & $67.1\pm1.7$ & $0.347\pm0.028$ \\
         \midrule
         \midrule
        \multirow{3}{*}{QAT-INT4 + DQ Masking} & GCN & $78.5\pm1.4$ & $62.8\pm8.5$ & $0.599\pm0.015$ \\
         & GAT & $64.4\pm9.3$ & $68.9\pm1.2$ & $0.529\pm0.008$ \\
         & GIN & $71.2\pm2.9$ & $56.7\pm3.8$ & $0.427\pm0.010$ \\
         \midrule
        \multirow{3}{*}{nQAT-INT4 + Percentile} & GCN & $75.6\pm2.5$ & $64.8\pm3.8$ & $0.633\pm0.012$ \\
         & GAT & $70.1\pm2.8$ & $51.4\pm3.4$ & $0.596\pm0.008$ \\
         & GIN & $63.5\pm2.0$ & $46.3\pm4.1$ & $0.771\pm0.058$ \\
        \bottomrule
    \end{tabular}
    \captionsetup{font=small,labelfont=bf}
    \caption{Ablation study against the two elements of Degree-Quant (DQ).
    The first two rows of results are obtained with only the stochastic element of Degree-Quant enabled for INT8 and INT4.
    Percentile-based quantization ranges are disabled in these experiments.
    The bottom row of results were obtained with noisy quantization (nQAT) at INT4 with the use of percentiles.
    DQ masking alone is often sufficient to achieve excellent results, but the addition of percentile-based range tracking can be beneficial to increase stability.
    We can see that using nQAT with percentiles is not sufficient to achieve results of the quality DQ provides.}
    \label{tab:stochDQ}
\end{table*}

\begin{table*}
\centering
\tablefontsize
\setlength{\tabcolsep}{4pt}

\begin{tabular}{@{}cc|ccc|ccc|ccc@{}}
\toprule
\multirow{2}{*}{Device} & \multirow{2}{*}{Arch.} & \multicolumn{3}{c}{CIFAR-10} & \multicolumn{3}{c}{Cora} & \multicolumn{3}{c}{Citeseer}\\
                        &                        & FP32 & W8A8 & Speedup & FP32 & W8A8 & Speedup & FP32 & W8A8 & Speedup \\ 
                        \midrule
\multirow{3}{*}{CPU}    & GCN                    & 182ms    & 88ms & $2.1\times$ & 0.94ms    &  0.74ms &  $1.3\times$  & 0.97ms    &  0.76ms &  $1.3\times$ \\
                        & GAT                    & 500ms    & 496ms & $1.0\times$ & 0.86ms    &  0.78ms &  $1.1\times$  & 0.99ms    &  0.88ms &  $1.1\times$ \\
                        & GIN                    & 144ms    & 44ms & $3.3\times$ & 0.85ms    &  0.68ms &  $1.3\times$  & 0.95ms    &  0.55ms &  $1.7\times$ \\
                        \midrule
\multirow{3}{*}{GPU}    & GCN                    & 2.1ms & 1.6ms & $1.3\times$ & 0.08ms    &  0.09ms &  $0.9\times$  & 0.09ms    &  0.09ms &  $1.0\times$ \\
                        & GAT                    & 30.0ms & 27.1ms & $1.1\times$ & 0.57ms    &  0.64ms &  $0.9\times$  & 0.56ms    &  0.64ms &  $0.9\times$ \\
                        & GIN                    & 20.9ms & 16.2ms  & 1.2$\times$ & 0.09ms    &  0.07ms &  $1.3\times$  & 0.09ms    &  0.07ms &  $1.3\times$ \\ 
                        \bottomrule
\end{tabular}

\captionsetup{font=small,labelfont=bf}
    \captionof{table}{INT8 latency results run on a 22 core 2.1GHz Intel Xeon Gold 6152 and, on a GTX 1080Ti GPU. All layers have 128 in/out features. For CIFAR-10 we used batch size of 1K graphs.}
    \label{tab:latencyCIFAR}
        
\end{table*}

\begin{figure}[h]
    \centering
    \includegraphics[width=\textwidth]{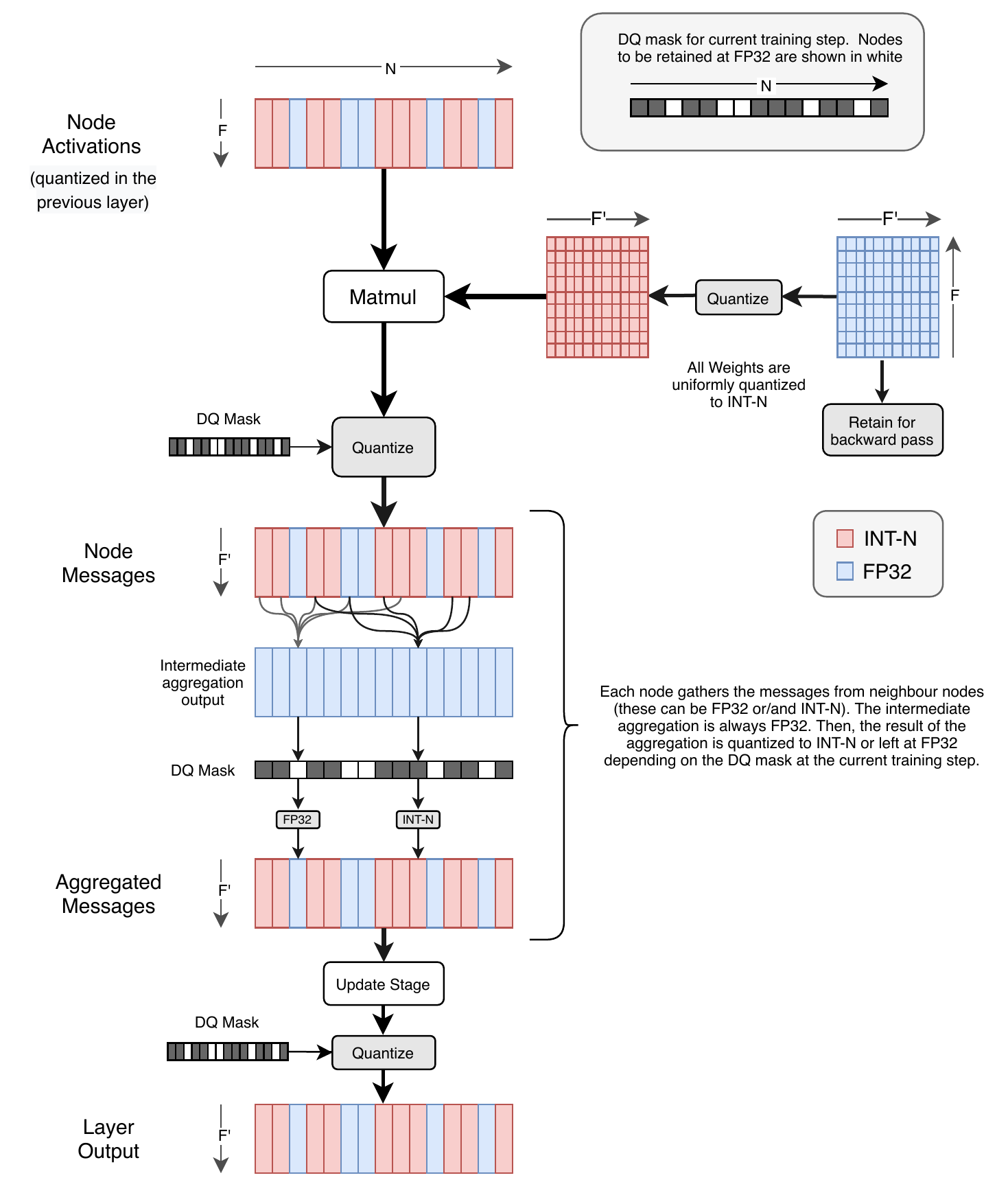}
    \captionsetup{font=small,labelfont=bf}
    \caption{Diagram representing how DQ makes use of a topology-aware quantization strategy that is better suited for GNNs. The diagram illustrates this for a GCN layer. At every training stage, a degree-based mask is generated. This mask is used in all quantization layers located after each of the stages in the message-passing pipeline. By retaining at FP32 nodes with higher-degree more often, the noisy updates during training have a lesser impact and therefore models perform better, even at INT4.}
    \label{fig:diagram_dq}
\end{figure}

\begin{figure}[h]
    \centering
    \includegraphics[width=\textwidth]{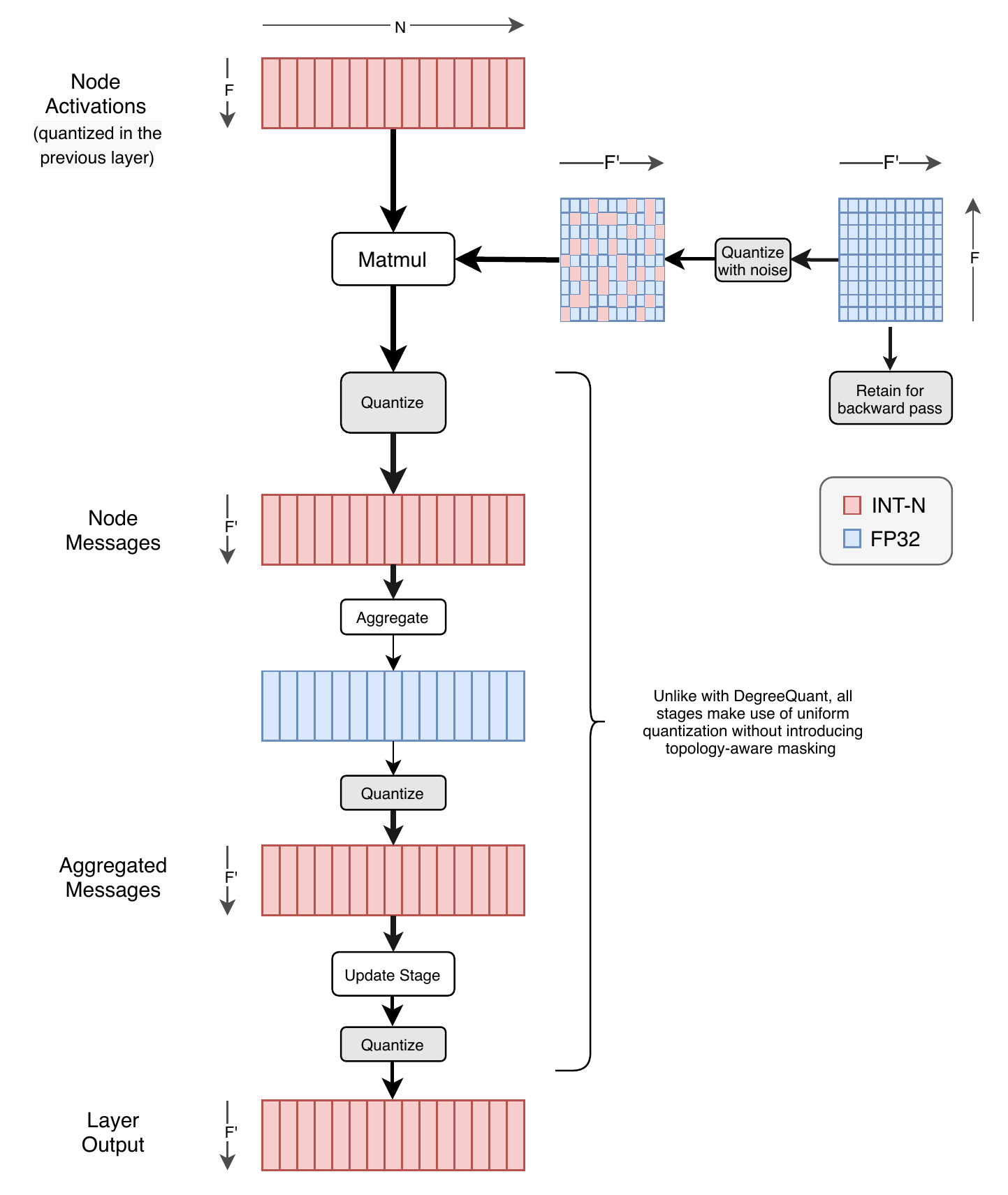}
    \captionsetup{font=small,labelfont=bf}
    \caption{Diagram representing how nQAT is implemented for GNNs. The diagram illustrates this for a GCN layer. The stochastic stage only takes place when quantizing the weights, the remaining of the quantization modules happen following a standard QAT strategy. A QAT diagram would be similar to this one but fully quantizing the weights.}
    \label{fig:diagram_qat}
\end{figure}

\begin{figure}[h]
    \centering
    \includegraphics[width=\textwidth]{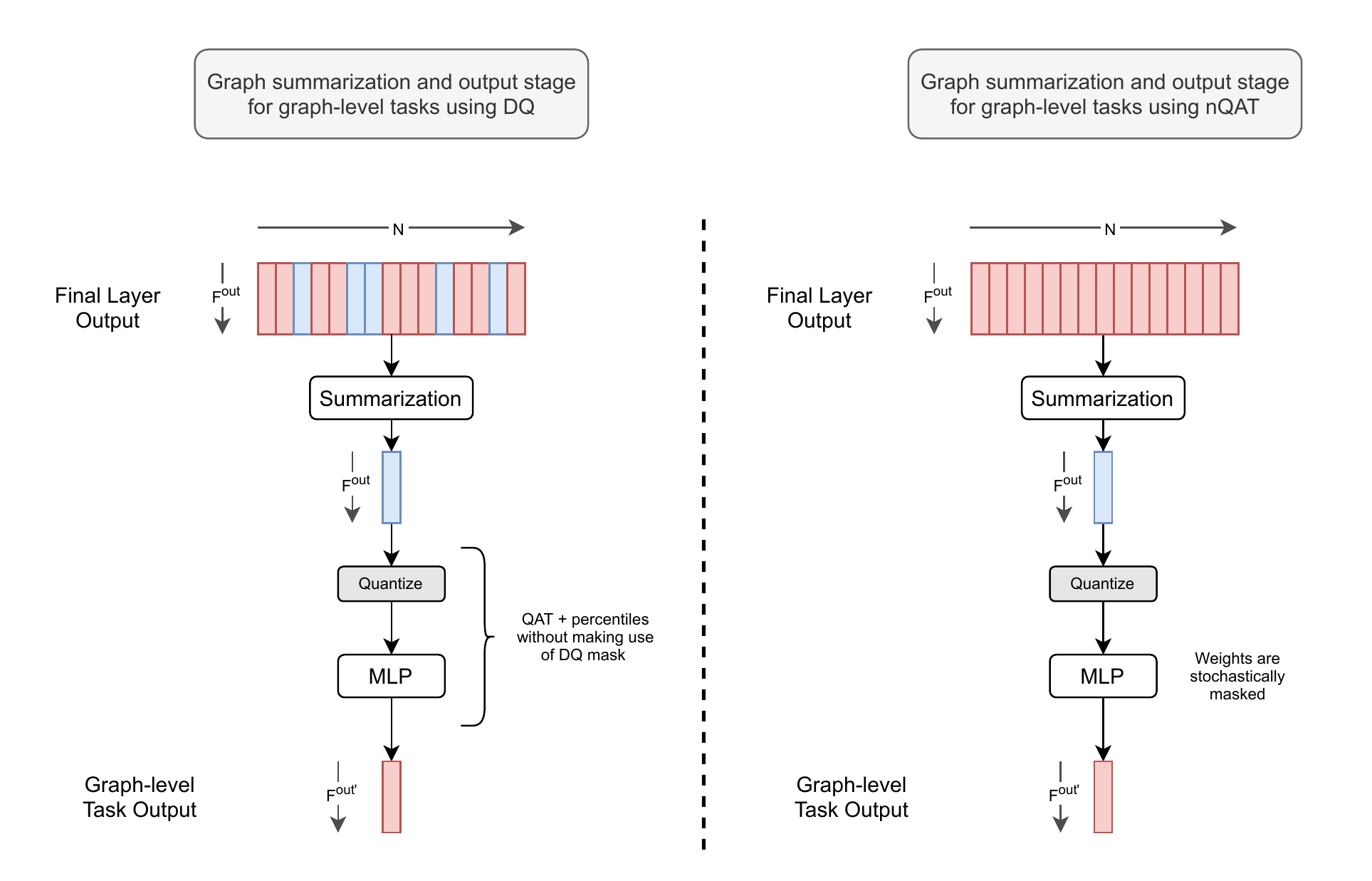}
    \captionsetup{font=small,labelfont=bf}
    \caption{Diagrams representing how the output graph-summarization stages for graph-level tasks (e.g. graph classification, graph regression) are implemented when making use of DQ (left) and nQAT (right). GNNs making use of DQ during the node-aggregation stages (see \cref{fig:diagram_dq}), do not use the stochastic element of DQ in the output MLP layers but still make use of percentiles. For models making use of nQAT, the final MLP still makes use of stochastic quantization of weights.}
    \label{fig:diagram_summary}
\end{figure}

\clearpage
\section{Code Listings}\label{sec:code}
Our code depends on PyTorch Geometric~\citep{fey2019fast}.
These snippets should be compatible with Python 3.7 and PyTorch Geometric version 1.4.3.
You can see the full code on GitHub: \url{https://github.com/camlsys/degree-quant}.

\subsection{Mask Generation}
\scriptsize
\begin{verbatim}
class ProbabilisticHighDegreeMask:
    def __init__(self, low_prob, high_prob, per_graph=True):
        self.low_prob = low_prob
        self.high_prob = high_prob
        self.per_graph = per_graph

    def _process_graph(self, graph):
        # Note that:
        # 1. The probability of being masked increases as the indegree increases
        # 2. All nodes with the same indegree have the same bernoulli p
        # 3. you can set this such that all nodes have some probability of being masked

        n = graph.num_nodes
        indegree = degree(graph.edge_index[1], n, dtype=torch.long)
        counts = torch.bincount(indegree)

        step_size = (self.high_prob - self.low_prob) / n
        indegree_ps = counts * step_size
        indegree_ps = torch.cumsum(indegree_ps, dim=0)
        indegree_ps += self.low_prob
        graph.prob_mask = indegree_ps[indegree]

        return graph

    def __call__(self, data):
        if self.per_graph and isinstance(data, Batch):
            graphs = data.to_data_list()
            processed = []
            for g in graphs:
                g = self._process_graph(g)
                processed.append(g)
            return Batch.from_data_list(processed)
        else:
            return self._process_graph(data)


def evaluate_prob_mask(data):
    return torch.bernoulli(data.prob_mask).to(torch.bool)
\end{verbatim}

\subsection{Message Passing with Degree-Quant}
\normalsize
Here we provide code to implement the layers as used by our proposal.
These are heavily based off of the classes provided by PyTorch Geometric, with only minor modifications to insert the quantization steps where necessary.
The normal quantized versions are similar, except without any concept of high/low masking.

\scriptsize
\begin{verbatim}
class MessagePassingMultiQuant(nn.Module):
    """This class is a lightweight modification of the default PyTorch
    Geometric MessagePassing class"""
    
    # irrelevant methods removed
    
    def propagate(self, edge_index, mask, size=None, **kwargs):
        # some lines skipped ...
        msg = self.message(**msg_kwargs)
        if self.training:
            # This is for the masking of messages:
            edge_mask = torch.index_select(mask, 0, edge_index[0])
            out = torch.empty_like(msg)
            out[edge_mask] = self.mp_quantizers["message_high"](msg[edge_mask])
            out[~edge_mask] = self.mp_quantizers["message_low"](
                msg[~edge_mask]
            )
        else:
            out = self.mp_quantizers["message_low"](msg)

        aggr_kwargs = self.__distribute__(self.__aggr_params__, kwargs)
        aggrs = self.aggregate(out, **aggr_kwargs)
        if self.training:
            out = torch.empty_like(aggrs)
            out[mask] = self.mp_quantizers["aggregate_high"](aggrs[mask])
            out[~mask] = self.mp_quantizers["aggregate_low"](aggrs[~mask])
        else:
            out = self.mp_quantizers["aggregate_low"](aggrs)

        update_kwargs = self.__distribute__(self.__update_params__, kwargs)
        updates = self.update(out, **update_kwargs)
        if self.training:
            out = torch.empty_like(updates)
            out[mask] = self.mp_quantizers["update_high"](updates[mask])
            out[~mask] = self.mp_quantizers["update_low"](updates[~mask])
        else:
            out = self.mp_quantizers["update_low"](updates)

        return out
\end{verbatim}

\subsubsection{GCN}

\begin{verbatim}
class GCNConvMultiQuant(MessagePassingMultiQuant):
    # Some methods missed...
    def forward(self, x, edge_index, mask, edge_weight=None):
        # quantizing input
        if self.training:
            x_q = torch.empty_like(x)
            x_q[mask] = self.layer_quantizers["inputs_high"](x[mask])
            x_q[~mask] = self.layer_quantizers["inputs_low"](x[~mask])
        else:
            x_q = self.layer_quantizers["inputs_low"](x)

        # quantizing layer weights
        w_q = self.layer_quantizers["weights_low"](self.weight)
        if self.training:
            x = torch.empty((x_q.shape[0], w_q.shape[1])).to(x_q.device)
            x_tmp = torch.matmul(x_q, w_q)
            x[mask] = self.layer_quantizers["features_high"](x_tmp[mask])
            x[~mask] = self.layer_quantizers["features_low"](x_tmp[~mask])
        else:
            x = self.layer_quantizers["features_low"](torch.matmul(x_q, w_q))
        
        if self.normalize:
            edge_index, norm = self.norm(
                edge_index,
                x.size(self.node_dim),
                edge_weight,
                self.improved,
                x.dtype,
            )
        else:
            norm = edge_weight
            
        norm = self.layer_quantizers["norm"](norm)
        return self.propagate(edge_index, x=x, norm=norm, mask=mask)
\end{verbatim}

\end{document}